\documentclass[10pt,twocolumn,letterpaper]{article}

\usepackage{cvpr}              

\usepackage{xcolor}

\definecolor{darkgreen}{rgb}{0.0, 0.5, 0.0}
\definecolor{olive}{rgb}{0.4, 0.4, 0.0}
\definecolor{darkred}{rgb}{0.8, 0.0, 0}

\definecolor{cvprblue}{rgb}{0.21,0.49,0.74}
\usepackage[pagebackref,breaklinks,colorlinks,allcolors=cvprblue]{hyperref}

\def\paperID{} 
\def\confName{CVPR}
\def\confYear{2026}

\title{SIGMA: Selective-Interleaved Generation with Multi-Attribute Tokens}

\author{
  Xiaoyan Zhang\textsuperscript{1,2}
  Zechen Bai\textsuperscript{4}
  Haofan Wang\textsuperscript{3}
  Yiren Song\textsuperscript{4}\textsuperscript{$\dagger$} \\ 
  \textsuperscript{1} Creatly AI \textsuperscript{2} University of Michigan  \textsuperscript{3} Lovart AI  \textsuperscript{4} National University of Singapore \\
}

\usepackage{multirow}
\usepackage[table]{xcolor}
\usepackage{booktabs}
\usepackage{subcaption}

\begin{document}



\twocolumn[{
\renewcommand\twocolumn[1][]{#1}
\maketitle
\vspace{-11mm}
\begin{center}
    \captionsetup{type=figure}
    \includegraphics[width=0.96\linewidth]{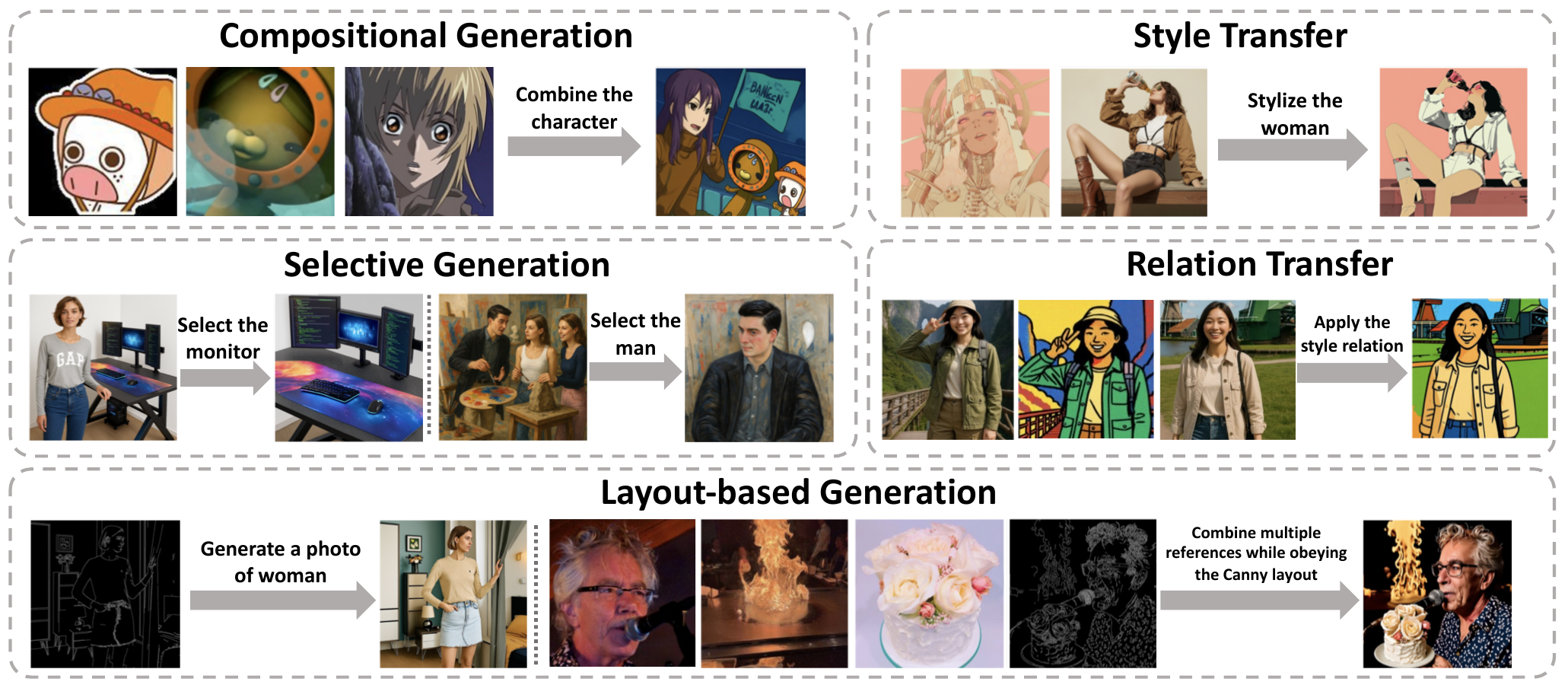}
    \vspace{-3mm}
    \captionof{figure}{Overview of tasks within our unified framework, covering diverse generation scenarios including compositional generation, selective generation, stylization, style relation transfer, and layout-based generation.}
    \label{fig:teaser}
\end{center}
}]

\begingroup
\renewcommand\thefootnote{}
\footnotetext{$\dagger$ Corresponding author.}
\endgroup

\begin{abstract}

Recent unified models such as Bagel demonstrate that paired image–edit data can effectively align multiple visual tasks within a single diffusion transformer. However, these models remain limited to single-condition inputs and lack the flexibility needed to synthesize results from multiple heterogeneous sources. We present SIGMA (Selective-Interleaved Generation with Multi-Attribute Tokens), a unified post-training framework that enables interleaved multi-condition generation within diffusion transformers. SIGMA introduces selective multi-attribute tokens, including style, content, subject and identity tokens, which allow the model to interpret and compose multiple visual conditions in an interleaved text–image sequence. Through post-training on the Bagel unified backbone with 700K interleaved examples, SIGMA supports compositional editing, selective attribute transfer and fine-grained multimodal alignment. Extensive experiments show that SIGMA improves controllability, cross-condition consistency and visual quality across diverse editing and generation tasks, with substantial gains over Bagel on compositional tasks. Code is available at \url{https://github.com/auihund/SIGMA}.

\vspace{-0.4cm}
\end{abstract}

\vspace{-0.5cm}
\section{Introduction}
\label{sec:intro}

Unified generative models~\cite{sun2023emu,tong2024metamorphmultimodalunderstandinggeneration,wu2025janus,xie2024show,team2024chameleon,huang2025wegen} have recently emerged as a powerful paradigm for combining diverse visual tasks within a single architecture.
The Bagel model~\cite{deng2025emergingpropertiesunifiedmultimodal} achieves impressive generalization in unified image generation and editing by training on paired image–edit data, allowing one diffusion transformer to perform both synthesis and manipulation.
However, such unified models are typically restricted to single-condition inputs—for example, a single reference image or prompt—thus limiting their ability to compose information from multiple sources.
In many practical scenarios, generation requires combining heterogeneous conditions, such as identity, content, and artistic style, into one coherent visual result.

This issue of binding~\cite{greff2020binding}, or how elements from different sources can be combined in a unified representation, has been a longstanding challenge in representation learning. In perceptual modeling, object-centric frameworks~\cite{locatello2020object,wu2022slotformer,lin2020space} address this problem by disentangling and tracking object-specific representations across scenes. However, in generative and editing settings, the binding problem reappears in a different form, where models must learn to associate semantic attributes such as identity, style, or layout with their visual targets. Earlier approaches~\cite{huang2017arbitrary,park2019semantic,lee2018diverse} typically achieved binding through task-specific architectural designs (e.g., using separate encoders for content and style) or by overfitting to a single editing modality. These methods often fail to generalize across diverse conditions, especially in auto-regressive architectures like transformers, which are key to models like Bagel~\cite{deng2025emergingpropertiesunifiedmultimodal}.

To address this limitation, we propose SIGMA (Selective-Interleaved Generation with Multi-Attribute Tokens), a post-training framework that extends the unified Bagel model to handle interleaved multi-condition generation.
SIGMA introduces a mechanism for binding that allows users to upload multiple condition images, such as a person photo, an accessory image, and a style reference, and describe their relationships through interleaved text–image sequences (e.g., “a photo of a man” + his portrait + “with a photo of a dog” + the dog image + “in the style of Van Gogh” + a style image).
The diffusion transformer then interprets this mixed sequence to produce a coherent, attribute-composed image, effectively bridging the gap between multi-reference input and unified generation.

At the core of SIGMA is the introduction of multi-attribute tokens, which enable the model to selectively control which aspects of each condition are used during generation.
We design specialized tokens for attributes such as style, content, identity, and subject, allowing fine-grained control over how different references contribute to the final output.
For instance, when a portrait of Van Gogh is encoded under a Style Token, the system extracts his artistic brushwork, while encoding it under an Identity Token transfers his facial identity.
This selective conditioning mechanism empowers SIGMA to perform attribute-specific reasoning and cross-modal composition, enabling a form of binding that is essential for flexible image editing and synthesis which are not supported by existing unified models.

We further perform interleaved post-training on the Bagel backbone using a newly collected dataset of 700K interleaved examples, spanning diverse combinations of reference images, textual prompts, and style–content mappings.
This post-training not only enhances compositional understanding but also allows flexible user interaction, supporting mixed text–image inputs and partial editing without fine-tuning.
Extensive experiments across multiple generation and editing benchmarks show that SIGMA substantially improves controllability, visual coherence, and attribute alignment. In particular, SIGMA achieves clear improvements over the Bagel unified model across compositional, selective, and layout-based generation, while approaching the performance of GPT–4o and Nano-Banana on several challenging benchmarks. An overview of the tasks supported by our unified framework is illustrated in Figure~\ref{fig:teaser}.

Our core contributions are:
\begin{itemize}
    \item We propose SIGMA, a unified post-training framework based on Bagel that supports interleaved multi-condition generation through a diffusion transformer architecture.
    \item We introduce a set of selective tokens that enable fine-grained control over how multiple condition images are composed and interleaved during generation.
    \item We construct a 700K interleaved image–text dataset and conduct extensive experiments, demonstrating SIGMA’s superior controllability, compositionality, and visual fidelity across various editing and synthesis tasks.
\end{itemize}

\section{Related Work}

\subsection{Image Generation}
\vspace{-1mm}
Recent advances in generative modeling have enhanced the fidelity and controllability of image synthesis. Early frameworks based on Generative Adversarial Networks (GANs)~\cite{goodfellow2020generative,karras2019style} demonstrated capabilities in generating high-resolution and realistic imagery, yet suffered from training instability and limited diversity. The introduction of Denoising Diffusion Models (DDMs)~\cite{ho2020denoising,rombach2022high} fundamentally reshaped the field, replacing adversarial training with iterative denoising. These models capture richer multi-scale visual statistics and enable fine-grained control through noise scheduling and classifier-free guidance. More recently, Diffusion Transformers (DiTs)~\cite{peebles2023scalable} unified diffusion dynamics with transformer-based architectures, demonstrating superior scalability and cross-domain generalization~\cite{song2025makeanything, song2025layertracer, huang2025arteditor, wang2025diffdecompose, li2025ic, liu2025omnipsd, zhang2025enhancing, feng2025dit4edit}. Building upon these foundations, subsequent works such as Flux~\cite{labs2025flux1kontextflowmatching}, PixArt-$\alpha$~\cite{chen2023pixart}, and OmniGen~\cite{xiao2025omnigen} explored cross-modal conditioning and large-scale pretraining to achieve both visual quality and semantic alignment.

\subsection{Conditional Generation}
\vspace{-1mm}
Conditioned image generation aims to control the generative process via external signals such as subject ~\cite{jiang2025personalized, song2024processpainter, guo2025any2anytryon, zhang2025markovian, zhu2024instantswap}, layout ~\cite{zhang2023adding, xie2023omnicontrol, zhang2025easycontrol, ma2024followpose, ma2024followyouremoji, ma2025followyourclick, ma2025followyourmotion},  font ~\cite{shi2024fonts, lu2025easytext, shi2025wordcon}, or style cues ~\cite{gong2025relationadapter, song2025omniconsistency, zhang2025magiccolor}. Text-to-image diffusion models, including Stable Diffusion~\cite{rombach2022high}, Imagen~\cite{saharia2022photorealistic}, and DALL·E 2~\cite{ramesh2022hierarchicaltextconditionalimagegeneration}, demonstrated unprecedented semantic controllability by aligning latent representations with language embeddings. Beyond textual prompts, multimodal conditioning strategies have emerged to guide generation via spatial and structural constraints such as ControlNet~\cite{zhang2023adding}, T2I-Adapter~\cite{mou2024t2i}, and IP-Adapter~\cite{ye2023ip}. These methods enable precise manipulation over pose, depth, or sketch inputs. However, existing pipelines often rely on independent modules or task-specific fine-tuning, limiting cross-condition generalization. 

\subsection{Unified Generative Models}
\vspace{-1mm}
While task-specific conditional models achieve high controllability, they typically lack cross-domain generalization and parameter efficiency. The emerging trend is to build unified generative frameworks that jointly support multiple conditioning modalities within a single backbone. Representative efforts include OmniControl~\cite{xie2023omnicontrol}, Make-A-Scene~\cite{gafni2022make}, and PixArt-$\Sigma$~\cite{chen2024pixart}, which extend diffusion transformers to handle multimodal prompts such as text, layout, and sketches in a unified latent space. Other approaches, such as UniDiffuser~\cite{bao2023one} and Muse~\cite{chang2023muse}, jointly learn bidirectional mappings between vision and language, supporting both generation and understanding within the same model ~\cite{ye2025loom}. These unified paradigms highlight a shift toward general-purpose generative intelligence—models capable of seamlessly adapting to diverse conditions and tasks without task-specific retraining. Our work follows this trend by designing a unified diffusion–transformer architecture that maintains modality-consistent representations while preserving fine-grained conditional controllability.

\section{Method}
\label{sec:method}

\begin{figure*}[t]
    \centering
    \includegraphics[width=\linewidth]{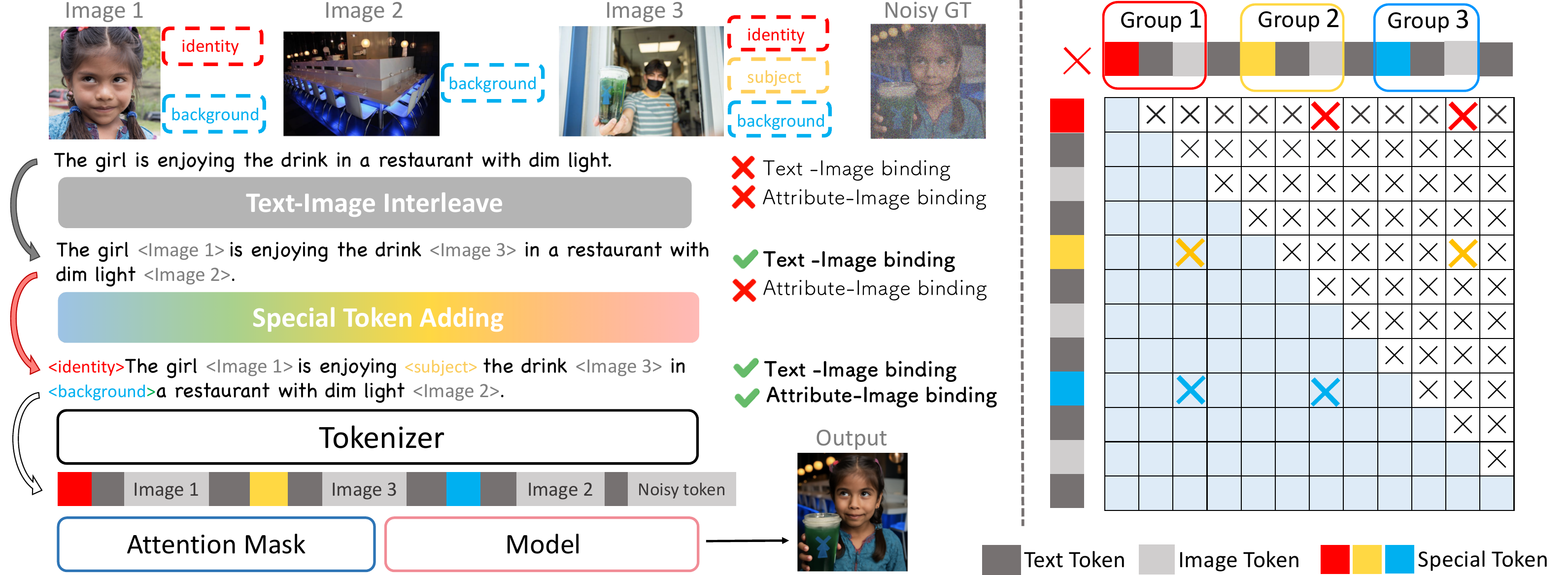}
    \caption{Overview of SIGMA. \textbf{Left:} Each sample may include multiple images, and each image can involve multiple visual attributes. Text–Image Interleave aligns textual spans with image placeholders, while Special Token Adding binds specific attributes to their corresponding images, avoiding semantic entanglement. \textbf{Right:} On top of causal attention, we add group-scoped masks so each special token can only attend to images within its own group, effectively reducing cross-image interference and ensuring clean attribute–image binding.}
    \label{fig:overview}
\end{figure*}

In this section, we describe the unified backbone and post-training setup in Sec.~\ref{sec:unified},the multi-attribute token design in Sec.~\ref{sec:token}, the interleaved conditioning mechanism in Sec.~\ref{sec:interleave}, and the group-scoped attention mask in Sec.~\ref{sec:mask}.  The overall architecture is illustrated in Fig.~\ref{fig:overview}.

\subsection{Unified Backbone and Post-Training}
\vspace{-0.5mm}
\label{sec:unified}

We build upon the unified model paradigm established by Bagel, which integrates multiple image generation and editing tasks within a single diffusion transformer backbone. 
Let $\mathcal{M}_{\theta}$ denote the pretrained unified diffusion model parameterized by $\theta$, trained on paired data $(x_{\text{src}}, x_{\text{tgt}})$ representing pre- and post-edit images. 
Given an input image $x_{\text{src}}$ and a text instruction $c$, the model predicts the target image $x_{\text{tgt}}$ through the denoising process:
\begingroup
\setlength{\abovedisplayskip}{4pt}
\setlength{\belowdisplayskip}{4pt}
\begin{equation}
    \mathbf{z}_{t-1} = \mathcal{M}_{\theta}(\mathbf{z}_t, c, x_{\text{src}}, t),
\end{equation}
where $\mathbf{z}_t$ denotes the latent at timestep $t$. 
This unified setup allows multi-task learning across generation, editing, and inpainting.
\endgroup

However, the original unified model is restricted to single-condition inputs and cannot handle multimodal conditioning (e.g., multiple reference images with different semantics). 
To overcome this limitation, we introduce a \textbf{post-training phase} that augments the Bagel backbone to support \textit{interleaved multi-condition inputs}. 
Instead of conditioning on a single pair $(x_{\text{src}}, c)$, SIGMA is trained on interleaved sequences $s$ that mix multiple reference images and text spans with multi-attribute tokens.
We keep the same denoising objective as in the backbone and simply change the conditioning form:
\vspace{-2mm}
\begin{equation}
    \mathcal{L}_{\text{SIGMA}}
    \;=\;
    \mathbb{E}_{(s, x_{\text{tgt}}),\,t}\big[
        \big\|
            \mathbf{z}_{t-1} - \mathcal{M}_{\theta}(\mathbf{z}_t, s, t)
        \big\|_2^2
    \big],
\end{equation}
where $s$ denotes the interleaved text–image sequence that includes the text prompt, all condition images, and their associated attribute tokens.

\subsection{Multi-Attribute Token Design}
\vspace{-0.5mm}
\label{sec:token}

The key innovation of SIGMA lies in its \textbf{Selective Multi-Attribute Tokenization}. 
Instead of treating all conditioning images equally, we assign a specific \textit{attribute token} to each condition according to its semantic role. 
Formally, for an input image $x_i$, we define a task-specific embedding $\tau_i \in \mathcal{T}$ chosen from a fixed attribute vocabulary:
\begin{equation}
    \mathcal{T} = \{\texttt{Style}, \texttt{Subject}, \texttt{Identity}, \texttt{Layout}, \dots\}.
\end{equation}
Each token $\tau_i$ modulates the feature extraction and fusion process by controlling which latent subspace of the diffusion transformer is activated. 
This design allows selective extraction of visual attributes: for instance, encoding a Van Gogh portrait under a \texttt{Style} token captures brushstroke patterns, while encoding it under an \texttt{Identity} token preserves facial characteristics.

Given the encoded condition image features $\mathbf{v}_i = E_{\phi}(x_i)$ and the corresponding attribute token $\tau_i$, we compute a token-conditioned embedding:
\begingroup
\setlength{\abovedisplayskip}{4pt}
\setlength{\belowdisplayskip}{4pt}
\begin{equation}
    \mathbf{t}_i = \mathbf{v}_i + W_{\tau}(\tau_i),
\end{equation}
where $W_{\tau}$ is a learnable attribute projection matrix. 
These attribute-specific tokens provide fine-grained control and facilitate multi-attribute composition during generation.
\endgroup

\subsection{Interleaved Conditioning Mechanism}
\label{sec:interleave}
\vspace{-0.5mm}
To support multi-condition fusion, we introduce an \textbf{interleaved conditioning} mechanism that allows text and image inputs to appear in an alternating sequence. 
Let $\mathbf{T}_k$ denote text embeddings and $\mathbf{I}_k$ denote the corresponding image condition embeddings with assigned attribute tokens. 
The final input sequence to the diffusion transformer is constructed as:
\begingroup
\setlength{\abovedisplayskip}{4pt}
\setlength{\belowdisplayskip}{4pt}
\begin{equation}
    \mathbf{H} = [\mathbf{T}_1; \mathbf{I}_1; \mathbf{T}_2; \mathbf{I}_2; \dots; \mathbf{T}_n; \mathbf{I}_n],
\end{equation}
\endgroup
where $[\cdot]$ denotes token concatenation in the temporal order of user specification. 
This interleaved structure enables flexible multimodal reasoning: the model can learn to parse textual descriptions and visual conditions jointly, preserving contextual alignment across modalities.


During training, the diffusion transformer processes this interleaved sequence through a series of cross-attention and self-attention layers. 
The Group-Scoped Attention Mask (Section~\ref{sec:mask}) ensures that only relevant attribute tokens influence specific attention heads, while implicit alignment feedback encourages consistent blending between adjacent conditions.
This alignment signal naturally arises from the denoising objective and the interleaved token structure, guiding the model to associate each attribute token with its corresponding visual source without requiring an explicit reward model.

At inference, users can freely combine different types of conditions (e.g., “a dog photo + a style painting + an environment image”) in any interleaved order. 
SIGMA interprets each condition adaptively based on its token semantics, producing coherent and controllable compositions across styles, identities, and layouts.

\subsection{Group-Scoped Attention Mask}
\label{sec:mask}
\vspace{-0.5mm}
Although interleaved conditioning enables flexible multimodal composition, it also introduces undesired \emph{attribute leakage}, where special tokens corresponding to one reference image may attend to unrelated image patches from other conditions.
Such uncontrolled attention often leads to semantic confusion between reference signals, resulting in degraded controllability and inconsistent generation. To address this, we design a simple but effective \textbf{group-scoped attention mask} that restricts cross-group attention for special tokens while preserving other attention patterns, thereby retaining the model’s ability to reason about spatial layouts, geometry, and alignment across conditions.

We denote the final input sequence by $\mathbf{H}=\{h_1,\dots,h_L\}$, where $L$ is the total number of tokens. Each token $h_\ell$ has a type $\mathrm{type}(h_\ell)\in\{\texttt{special},\texttt{text},\texttt{image},\texttt{plain}\}$, representing attribute tokens (e.g., \texttt{<id>}, \texttt{<style>}), textual instruction tokens, image patch tokens, and other non-grouped text tokens, respectively. Tokens of type \texttt{special} or \texttt{image} are additionally associated with a group index $\mathrm{grp}(h_\ell)\in\{1,\dots,m\}$, corresponding to each reference condition. We construct a binary attention mask $\mathbf{B}\in\{0,1\}^{L\times L}$, where each element $\mathbf{B}[q,k]$ determines whether the query token $h_q$ is allowed to attend to the key token $h_k$. The final mask is obtained by combining three components:
\begingroup
\setlength{\abovedisplayskip}{4pt}
\setlength{\belowdisplayskip}{4pt}
\begin{equation}
\mathbf{B} = (\mathbf{C}\wedge \mathbf{M}) \;\vee\; \mathbf{S},\qquad
\mathbf{A} = (\mathbf{1}-\mathbf{B})\cdot(-\infty),
\end{equation}
\endgroup

where $\mathbf{A}$ is added to the attention logits before the softmax. Here $\mathbf{C}$ encodes the causal attention pattern inherited from Bagel, $\mathbf{S}$ enables unrestricted intra-image attention, and $\mathbf{M}$ enforces the group constraint.

The causal mask $\mathbf{C}$ maintains the autoregressive property of the original model by allowing each token to attend only to previous tokens in the sequence:
\begingroup
\setlength{\abovedisplayskip}{4pt}
\setlength{\belowdisplayskip}{4pt}
\begin{equation}
\mathbf{C}[q,k]=1 \quad \text{iff} \quad k\le q.
\end{equation}
\endgroup
The intra-image mask $\mathbf{S}$ enables full bidirectional attention among patch tokens belonging to the same image, which is crucial for local reasoning about structure, geometry, and spatial relations:
\begingroup
\setlength{\abovedisplayskip}{4pt}
\setlength{\belowdisplayskip}{4pt}
\begin{equation}
\mathbf{S}[q,k] =
\begin{cases}
1, & \text{if } 
  \mathrm{type}(h_q)=\mathrm{type}(h_k)=\texttt{image}, \\
  & \quad \mathrm{img}(h_q)=\mathrm{img}(h_k), \\[3pt]
0, & \text{otherwise}.
\end{cases}
\end{equation}
\endgroup
Finally, the group constraint $\mathbf{M}$ restricts special tokens from attending to image patches outside their corresponding group, effectively blocking cross-condition leakage:
\begingroup
\setlength{\abovedisplayskip}{4pt}
\setlength{\belowdisplayskip}{4pt}
\begin{equation}
\mathbf{M}[q,k] =
\begin{cases}
0, & \mathrm{type}(h_q)=\texttt{special},\\
   & \mathrm{type}(h_k)=\texttt{image},\\
   & \mathrm{grp}(h_q)\neq \mathrm{grp}(h_k),\\[3pt]
1, & \text{otherwise}.
\end{cases}
\end{equation}
\endgroup

This masking strategy enforces minimal but effective structure in the attention map. Special tokens are linked only to their designated image patches. Images preserve unrestricted internal connectivity, and the rest of the sequence follows the original causal ordering. The design prevents attention drift across unrelated conditions while still allowing the model to capture global dependencies and relational cues through text tokens and causal connections. In practice, this yields attribute disentanglement and more controllable multi-condition generation without reducing the expressive capacity of the underlying diffusion transformer.

\section{Interleaved Multi-Condition Dataset}
\label{sec:dataset}
\vspace{-1mm}

\begin{figure*}[t]
    \centering
    \includegraphics[width=\linewidth]{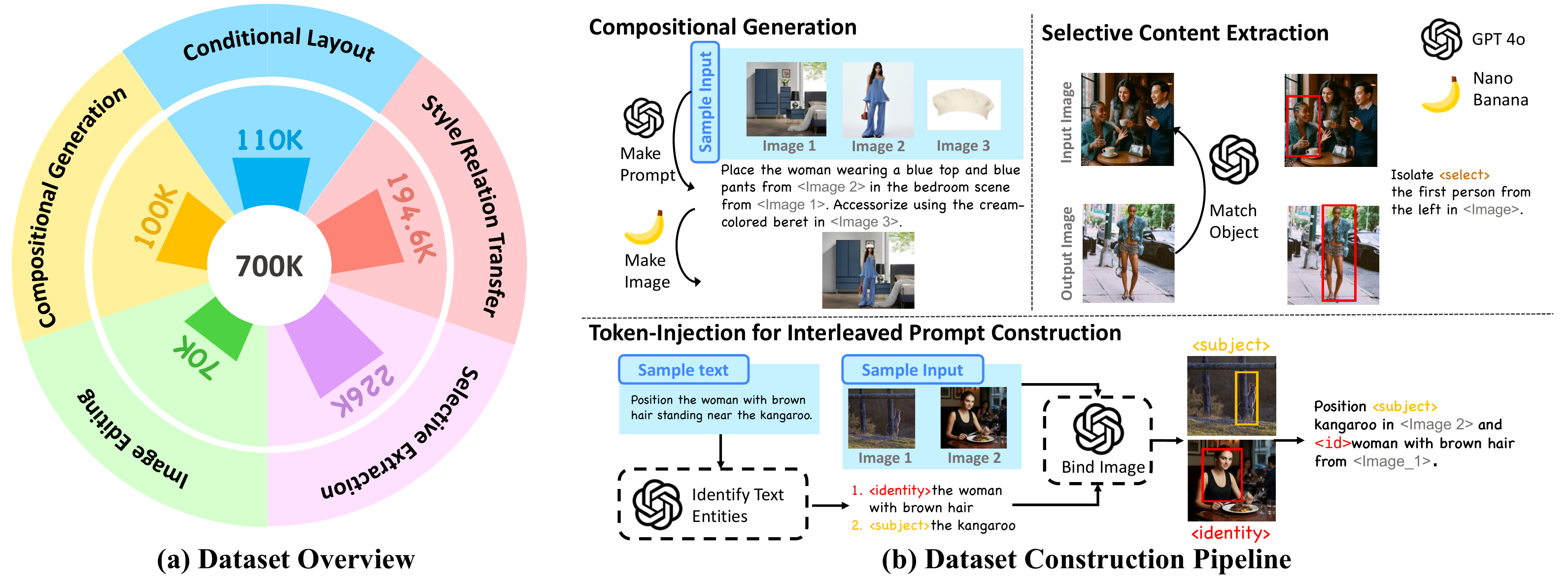}
    \vspace{-8mm}
    \caption{Overview and construction of the interleaved multi-condition dataset.\textbf{ (a)} The 700K corpus spans six task families, including compositional generation, selective content extraction, stylization, relation transfer, image editing, and conditional layout.
    \textbf{(b)} Data are built via compositional generation (GPT-4o + Nano-Banana), selective content extraction (object matching with GPT-4o), and token-injection that binds text entities with reference images for interleaved supervision.}
    \label{fig:data_overview}
\end{figure*}

\subsection{Dataset Overview}
\vspace{-1mm}
\label{subsec:dataset_overview}

To enable SIGMA to learn which attribute should be taken from which reference, we curate a large-scale interleaved dataset where text spans, special tokens, and multiple condition images are aligned. The final corpus contains \textbf{700K} sequences across six major task families—compositional generation (100K), selective content extraction (226K), stylization (153K), relation transfer (41.6K), image editing (70K), and conditional layout generation (110K). The distribution of these task families is illustrated in Fig.~\ref{fig:data_overview}(a). Compared with standard caption–image pairs, this interleaved formulation forces the model to handle multi-image, multi-attribute, and asymmetric-reference scenarios, which are central to controllable multimodal generation.

To make attribute–image bindings machine-readable, we introduce 14 special tokens denoting entity- or attribute-level cues such as identity, subject, clothing, style, layout, pose, and lighting. Tokens are injected directly before the corresponding textual mention, followed by the referenced image. This converts each caption into a structured multimodal sequence that specifies  which visual factor should be extracted from each source. Many samples are intentionally attribute-dense (e.g., \texttt{<subject>} + \texttt{<clothing>} + \texttt{<background>} for a single image), allowing SIGMA to learn disentangled selection rather than uncontrolled fusion when references overlap in content or appearance.

\subsection{Dataset Construction}
\vspace{-1mm}
\label{sec:dataset_construction}


Our dataset combines newly generated samples and adapted high-quality corpora.
We synthesize large-scale compositional, stylization, relational style-transfer, and editing data using GPT-4o and Nano-Banana, covering human, object, and scene combinations. The selective extraction subset is derived from Echo-4o by treating each compositional output as input and identifying extraction targets using GPT-4o. The conditional-layout subset incorporates canny/depth maps and layout-designated reference images, with geometric cues extracted by MiDaS. Existing datasets such as Nano-150K, Echo-4o, X2Edit, and ShareGPT-4o are converted into the interleaved format via token injection.

To unify data from heterogeneous sources, we employ a structured token-injection pipeline that converts each sample into an interleaved text–image sequence, as shown in Fig.~\ref{fig:data_overview}(b). Each entity-bearing phrase in the caption is paired with a special token and its corresponding reference image, yielding locally bound text–image groups that specify the intended visual attributes and their sources. This standardized interleaving process makes cross-condition relationships explicit and enables SIGMA to learn fine-grained, source-aware attribute control.

\section{Experiment}
\label{sec:exp}
\vspace{-1mm}

\subsection{Experimental Setup}
\vspace{-1mm}

\paragraph{Training Details.}

We build our model upon the Bagel unified diffusion backbone~\cite{deng2025emergingpropertiesunifiedmultimodal}, training only the generation branch while freezing the VAE module. For each task family in our interleaved multi-condition corpus, we allocate 95\% of the sequences to the training set, where multiple \emph{special tokens} serve as conditioning signals. The model is trained for 50K steps on 4 NVIDIA H200 GPUs. To improve efficiency, we apply token packing with a maximum of 30K tokens per packed batch. A cosine learning rate schedule is used with a peak learning rate of $2\times 10^{-5}$ and a minimum learning rate of $10^{-7}$. Optimization is performed using AdamW ($\beta_1=0.9$, $\beta_2=0.95$) with gradient clipping of $1.0$, and we employ fully sharded data parallelism to scale training across multiple GPUs.

\paragraph{Benchmarks.}
We evaluate our method on several benchmarks that cover different aspects of controllable image generation. We use XVerseBench~\cite{chen2025xverse} for compositional generation, which provides diverse human identities, objects, animals references for assessing multi-entity reasoning. In addition, we construct a new \textbf{comprehensive benchmark} that jointly evaluates controllability, compositional reasoning and structural alignment across four representative tasks: compositional generation, selective generation, style transfer and layout-based generation. All samples are drawn from the held-out portion of our corpus to ensure a strict separation between training and evaluation. Full statistics, benchmark construction details and per-task configurations are provided in the supplementary material.

\paragraph{Baselines.}

In addition to \textbf{Bagel}~\cite{deng2025emergingpropertiesunifiedmultimodal}, we include recent unified diffusion transformers such as \textbf{XVerse}~\cite{chen2025xverse} and \textbf{SSR}~\cite{zhang2024ssr}.
We also incorporate \textbf{EasyControl} for the layout based generation setting.
To provide a reference to general-purpose multimodal systems, we report results from closed-source models \textbf{GPT-4o}~\cite{openai2024gpt4ocard} and \textbf{Nano-Banana}~\cite{comanici2025gemini25pushingfrontier}.

\paragraph{Metrics.}
We evaluate controllability and perceptual quality using a mix of semantic and visual metrics.
\textbf{CLIP}$\uparrow$, \textbf{CLIP-I}$\uparrow$~\cite{radford2021learning}, and \textbf{DINO}$\uparrow$~\cite{oquab2023dinov2} similarities measure text–image and subject alignment, while \textbf{CLIP-ES}$\downarrow$ assesses subject exclusivity in multi-reference settings.
\textbf{AES}$\uparrow$~\cite{schuhmann2022laion} evaluates the overall aesthetic appeal, and \textbf{DreamSim}$\uparrow$~\cite{fu2023dreamsim} quantifies perceptual similarity aligned with human visual judgment.
For layout-based generation tasks, we additionally report the \textbf{F1}$\uparrow$ score computed between the extracted and input edge maps in edge-conditioned generation to assess structural consistency, and the \textbf{FID}$\downarrow$ to measure distributional fidelity.

\begin{figure*}[t]
    \centering
    \includegraphics[width=1.0\linewidth]{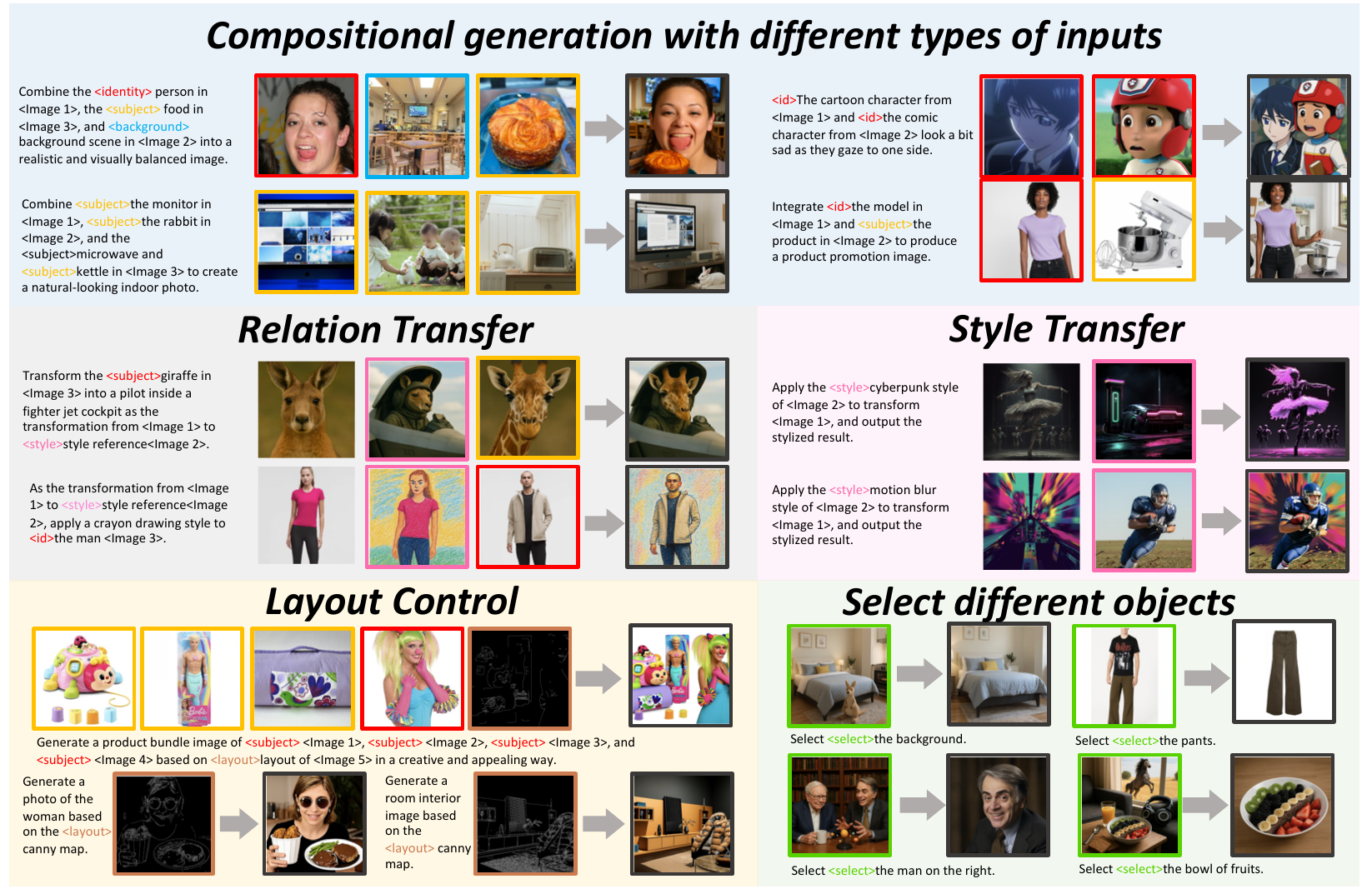}
    \vspace{-2mm}
    \caption{Qualitative results achieved by SIGMA. By leveraging specialized attribute tokens, SIGMA flexibly binds the required elements from input references, accomplishing a wide range of generation tasks. The results demonstrate clear binding between inputs and outputs, as well as high-quality, visually coherent generations across all scenarios.}
    \label{fig:results}
\end{figure*}

\begin{table}[t]
\centering
\footnotesize
\setlength{\tabcolsep}{3pt}
\renewcommand{\arraystretch}{1.05}
\resizebox{\linewidth}{!}{
\begin{tabular}{l l c c c c}
\toprule
\textbf{Benchmark} & \textbf{Method} &
\textbf{CLIP}$\uparrow$ &
\textbf{CLIP-I}$\uparrow$ &
\textbf{DINO}$\uparrow$ &
\textbf{DreamSim}$\uparrow$ \\
\midrule

\multirow{6}{*}{\textbf{XVerse Bench}}
  & \cellcolor{gray!12} GPT-4o
    & \cellcolor{gray!12} 32.94
    & \cellcolor{gray!12} \textbf{77.74}
    & \cellcolor{gray!12} \textbf{66.42}
    & \cellcolor{gray!12} 68.11 \\
  & \cellcolor{gray!12} Nano-Banana
    & \cellcolor{gray!12} 32.45
    & \cellcolor{gray!12} 75.50
    & \cellcolor{gray!12} 65.10
    & \cellcolor{gray!12} \textbf{69.54} \\
  & SSR
    & 27.72 & 69.19 & 46.98 & 63.24 \\
  & XVerse
    & \textbf{33.94} & 67.53 & 45.24 & 66.25 \\
  & Bagel
    & 24.32 & 66.32 & 56.13 & 53.31 \\
  & SIGMA (Ours)
    & 31.96 {\scriptsize(+7.64)} &
      75.57 {\scriptsize(+9.25)} &
      59.52 {\scriptsize(+3.39)} &
      67.87 {\scriptsize(+14.56)} \\
\midrule

\multirow{6}{*}{\textbf{Our Bench}}
  & \cellcolor{gray!12} GPT-4o
    & \cellcolor{gray!12} \textbf{31.07}
    & \cellcolor{gray!12} 77.93
    & \cellcolor{gray!12} 63.58
    & \cellcolor{gray!12} \textbf{67.49} \\
  & \cellcolor{gray!12} Nano-Banana
    & \cellcolor{gray!12} 30.65
    & \cellcolor{gray!12} 75.63
    & \cellcolor{gray!12} 62.22
    & \cellcolor{gray!12} 62.78 \\
  & SSR
    & 28.61 & 65.34 & 54.11 & 52.65 \\
  & XVerse
    & 32.33 & 44.15 & 42.76 & 54.63 \\
  & Bagel
    & 17.91 & 52.52 & 41.62 & 43.27 \\
  & \textbf{SIGMA (Ours)}
    & 30.29 {\scriptsize(+12.38)} &
      \textbf{78.94} {\scriptsize(+26.42)} &
      \textbf{64.08} {\scriptsize(+22.46)} &
      62.45 {\scriptsize(+19.18)} \\
\bottomrule
\end{tabular}}
\vspace{-2mm}
\caption{
Comparison of compositional generation performance across two benchmarks.
Numbers in parentheses show improvement over Bagel.
Rows highlighted in \textcolor{gray!80}{gray} denote closed-source models.
}
\label{tab:xverse_final2}
\vspace{-0.4cm}
\end{table}

\begin{table}[t]
\centering
\footnotesize
\setlength{\tabcolsep}{6pt}
\renewcommand{\arraystretch}{1.2}
\resizebox{0.95\linewidth}{!}{%
\begin{tabular}{lcccc}
\toprule
\textbf{Method} &
\textbf{CLIP}$\uparrow$ &
\textbf{CLIP-I}$\uparrow$ &
\textbf{CLIP-ES}$\downarrow$ &
\textbf{AES}$\uparrow$ \\
\midrule
\cellcolor{gray!12} GPT\textminus4o
    & \cellcolor{gray!12} 25.84
    & \cellcolor{gray!12} 80.14
    & \cellcolor{gray!12} 60.22
    & \cellcolor{gray!12} 5.882 \\
\cellcolor{gray!12} Nano\textminus Banana
    & \cellcolor{gray!12} 25.48
    & \cellcolor{gray!12} 77.88
    & \cellcolor{gray!12} 62.63
    & \cellcolor{gray!12} 5.804 \\
SSR             & 25.46 & 71.68 & 58.78 & \textbf{5.914} \\
XVerse          & 25.15 & 70.61 & 61.16 & 5.181 \\
Bagel           & 23.49 & 70.61 & 67.90 & 5.209 \\
\textbf{SIGMA (Ours)}
    & \textbf{25.90} {\scriptsize(+2.41)}
    & \textbf{80.26} {\scriptsize(+9.65)}
    & \textbf{58.02} {\scriptsize(–9.88)}
    & 5.849 {\scriptsize(+0.64)} \\
\bottomrule
\end{tabular}}
\vspace{-3mm}
\caption{Selective generation results on our benchmark. Numbers in parentheses show improvement over Bagel. }
\label{tab:selective_generation}
\vspace{-0.3cm}
\end{table}

\begin{table}[t]
\centering
\vspace{0.5em}
\resizebox{\linewidth}{!}{
\begin{tabular}{l l c c c c}
\toprule
\textbf{Condition} & \textbf{Method} & \textbf{F1}$\uparrow$ & \textbf{FID}$\downarrow$ & \textbf{CLIP}$\uparrow$ & \textbf{AES}$\uparrow$ \\
\midrule

\multirow{5}{*}{\textbf{Layout only}}
 & \cellcolor{gray!12} GPT\textminus4o 
    & \cellcolor{gray!12} 0.09 
    & \cellcolor{gray!12} 196.02 
    & \cellcolor{gray!12} 27.33 
    & \cellcolor{gray!12} 5.545 \\
 & \cellcolor{gray!12} Nano\textminus Banana 
    & \cellcolor{gray!12} 0.12 
    & \cellcolor{gray!12} 189.46 
    & \cellcolor{gray!12} 26.50 
    & \cellcolor{gray!12} \textbf{5.743} \\
 & EasyControl & 0.16 & 135.27 & 27.14 & 5.357 \\
 & Bagel & 0.10 & \textbf{103.72} & 25.82 & 4.13 \\
 & \textbf{SIGMA (Ours)} 
    & \textbf{0.44} {\scriptsize(+0.34)}
    & 121.08 {\scriptsize(+17.36)}
    & 26.35 {\scriptsize(+0.53)}
    & 5.649 {\scriptsize(+1.52)} \\
\midrule

\multirow{5}{*}{\shortstack{\textbf{Layout}\\\textbf{+ Reference}}}
 & \cellcolor{gray!12} GPT\textminus4o 
    & \cellcolor{gray!12} 0.04 
    & \cellcolor{gray!12} 182.48 
    & \cellcolor{gray!12} 24.39 
    & \cellcolor{gray!12} \textbf{6.01} \\
 & \cellcolor{gray!12} Nano\textminus Banana 
    & \cellcolor{gray!12} 0.04 
    & \cellcolor{gray!12} 161.91 
    & \cellcolor{gray!12} 24.86 
    & \cellcolor{gray!12} 5.912 \\
 & Bagel & 0.25 & 188.83 & 24.22 & 4.54 \\
 & \textbf{SIGMA (Ours)} 
    & \textbf{0.35} {\scriptsize(+0.10)}
    & \textbf{108.05} {\scriptsize(–80.78)}
    & \textbf{24.53} {\scriptsize(+0.31)}
    & 5.774 {\scriptsize(+1.23)} \\
\bottomrule
\end{tabular}}
\vspace{-0.2cm}
\caption{
Layout-based generation results on the layout-based subset of our benchmark.
Numbers in parentheses show improvement over Bagel. 
}
\label{tab:layout}
\end{table}

\begin{figure}[t]
    \centering
    \includegraphics[width=\linewidth]{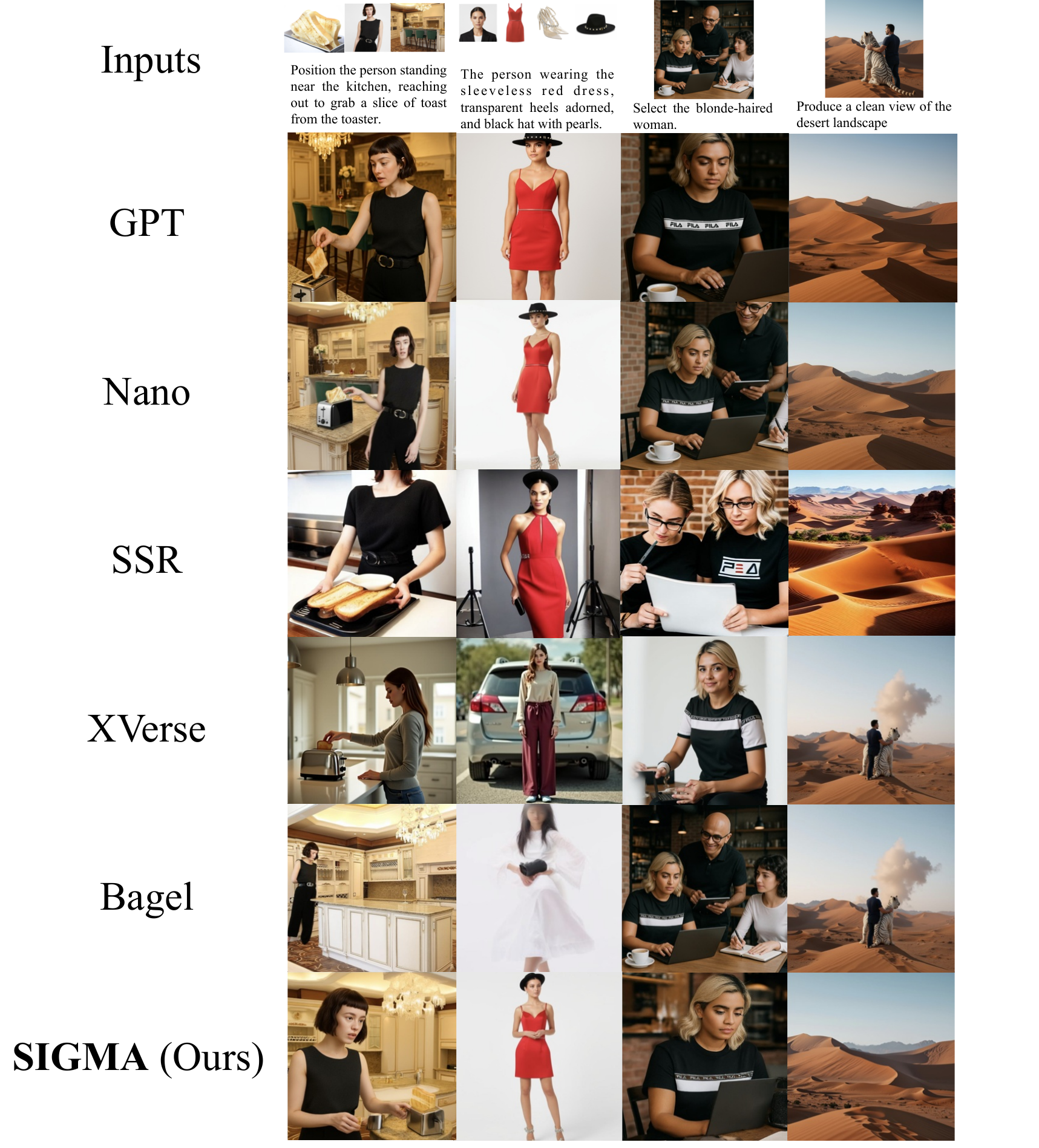}
    \vspace{-7mm}
    \caption{Qualitative comparisons on our benchmark.}
    \label{fig:comparison}
\end{figure}

\subsection{Quantitative Evaluation}
We assess SIGMA on three representative tasks. Across all settings, SIGMA delivers strong and reliable performance.

On the compositional generation benchmark, Table~\ref{tab:xverse_final2} shows that SIGMA outperforms both SSR and Bagel across all four metrics. Relative to Bagel, SIGMA achieves substantial gains on CLIP, CLIP-I, DINO, and DreamSim, highlighting the effectiveness of our fusion mechanism in preserving identity and compositional integrity. Compared with XVerse, SIGMA attains slightly lower CLIP scores but consistently higher CLIP-I, DINO, and DreamSim, indicating stronger structural and perceptual alignment with the references. Among all open-source methods, SIGMA delivers the best overall performance and approaches the quality of GPT–4o and Nano-Banana, suggesting that it can accurately capture the intended objects and attributes while maintaining geometric fidelity and fine-grained visual detail.
For selective generation, Table~\ref{tab:selective_generation} highlights SIGMA’s ability to pick out the correct target while maintaining visually pleasing results. The method obtains the lowest CLIP-ES score among all baselines, meaning it is less likely to select irrelevant objects. Together with its competitive CLIP and CLIP-I values, these results show that SIGMA can selectively extract and transform the desired object with dependable precision. 
For layout based generation, Table~\ref{tab:layout} shows that SIGMA follows layout constraints more accurately than prior unified models. The layout plus reference setting further demonstrates that SIGMA can align spatial structure with appearance guidance in a coherent manner.

\subsection{Qualitative Evaluation}

Figure~\ref{fig:results} presents an overview of the visual capabilities enabled by SIGMA. The model produces high-quality results across a series of tasks and remains stable even in challenging multi-input scenarios. SIGMA can merge heterogeneous references, follow complex textual descriptions, and generate images that preserve fine-grained details such as texture, lighting, and identity. The outputs are visually coherent and maintain consistent composition across difficult cases, demonstrating that our multi-condition design provides reliable and flexible control over how different reference elements are incorporated into the final generation.

To further assess practical behavior, Figure~\ref{fig:comparison} compares SIGMA with existing unified diffusion transformers. The examples illustrate performance under two demanding settings: compositional generation and selective generation. In the compositional setting, SIGMA can identify the correct objects from multiple visual references, integrate them naturally, and maintain stable geometrical relations and plausible scene structure. The generated images remain faithful to both the content and the text instructions. In comparison, SSR, XVerse, and Bagel often exhibit consistency issues, such as difficulty selecting the correct attributes and applying them properly during generation, while Nano maintains stronger consistency but sometimes produces unnatural shapes.
The selective generation examples highlight another strength of the model. Each input image contains multiple candidate objects, yet SIGMA reliably follows the user instruction and extracts the intended target, whether it is a specific object or a background. The selected elements appear natural and well-formed, preserving identity and appearance without artifacts. At the same time, irrelevant objects are effectively excluded, indicating strong exclusivity and consistent instruction following.

Overall, the qualitative results show that SIGMA handles diverse tasks in a unified framework while producing coherent, detailed, and semantically aligned generations.


\subsection{Ablation Study}
\vspace{-2mm}

\begin{table}[ht]
\centering
\small
\setlength{\tabcolsep}{6pt}
\renewcommand{\arraystretch}{1.2}
\resizebox{\linewidth}{!}{
\begin{tabular}{ccc|cccc}
\toprule
\multirow{1}{*}{\textbf{Special}} & \multirow{1}{*}{\textbf{Mask}} & \multirow{1}{*}{\textbf{All}} 
& \multicolumn{1}{c}{\textbf{CLIP}$\uparrow$} 
& \multicolumn{1}{c}{\textbf{CLIP-I}$\uparrow$} 
& \multicolumn{1}{c}{\textbf{DreamSim}$\uparrow$} 
& \multicolumn{1}{c}{\textbf{AES}$\uparrow$} \\
\midrule
 &  & \checkmark & 25.85 & 62.67 & 44.74 & 5.576 \\
\checkmark &  & \checkmark & 29.25 & 74.26 & 57.11 & 5.561 \\
\checkmark & \checkmark &  & 27.32 & 72.64 & 58.65 & 5.506 \\
\checkmark & \checkmark & \checkmark & \textbf{30.29} & \textbf{78.94} & \textbf{62.45} & \textbf{5.731} \\
\bottomrule
\end{tabular}}
\vspace{-3mm}
\caption{Ablation of SIGMA components.
``All" indicates full-parameter finetuning, while rows without a check in the ``All" column correspond to LoRA-based tuning.  
We analyze the contributions of special tokens and group-scoped attention masks under both setups.  
Higher values indicate better alignment or perceptual quality.}
\label{tab:abl_sigma_full}
\end{table}

We evaluate the contribution of multi-attribute tokens, the group-scoped attention mask, and the choice between full-parameter finetuning and LoRA tuning. As shown in Table~\ref{tab:abl_sigma_full}, removing attribute tokens leads to a substantial drop in CLIP and CLIP-I, indicating that the model can no longer reliably determine which attributes should be extracted or where they should be applied. Adding the group-scoped mask yields further gains in CLIP-I and DreamSim, reflecting improved structural and perceptual consistency. LoRA-based tuning remains functional but trails full finetuning in both alignment and aesthetic quality, suggesting that limited parameter updates reduce the model’s expressive capacity.

\begin{figure}[t]
  \centering
    \includegraphics[width=\columnwidth]{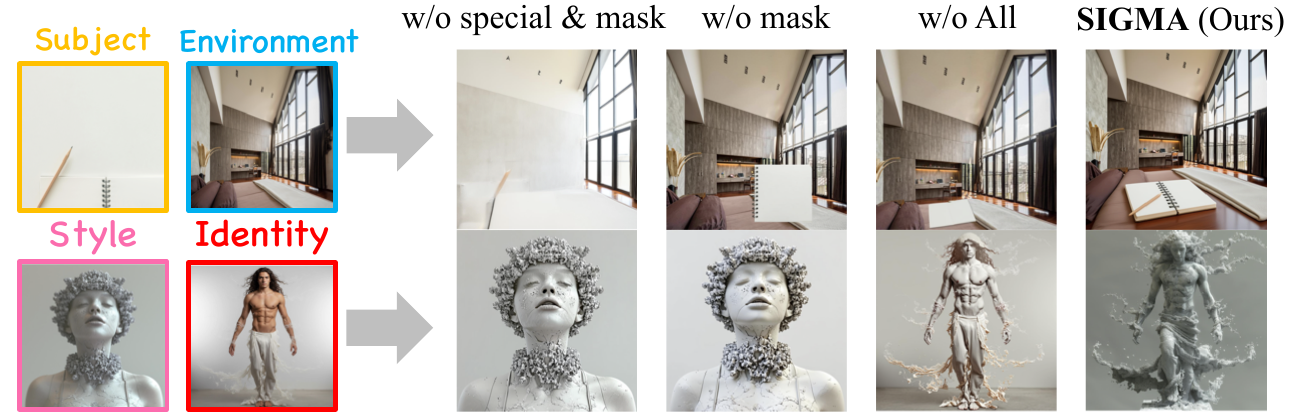}
    \vspace{-7mm}
    \caption{Qualitative ablation on attribute tokens, masked attention, and full-parameter training strategy.}
    \label{fig:ablation}
\end{figure}

Figure~\ref{fig:ablation} provides the qualitative results. In the compositional generation example, omitting attribute tokens causes the model to confuse object roles entirely, while adding them without masking produces clearer objects but still unnatural interactions. LoRA tuning generates reasonable layouts but misses fine details such as missing cables. In the stylization case, the absence of the mask causes the model to collapse toward copying the style reference, showing that it fails to bind the style token to the content image. When the mask is included, attribute binding becomes correct and style transfer is coherent. LoRA again produces correct but less consistent stylization, whereas SIGMA achieves the highest fidelity across all elements.

\section{Conclusion}
\label{sec:conclusion}

We presented SIGMA, a unified diffusion transformer for controllable multi-condition image generation. The framework combines selective multi-attribute tokens that explicitly specify what to extract from each reference image, an interleaved text and image conditioning scheme that enables flexible multi-reference fusion, and a group-scoped attention mask that reduces unwanted interactions across different sources. These components are trained on a 700K interleaved dataset covering compositional generation, selective content extraction, stylization, relation transfer and layout-guided synthesis. This large and diverse corpus helps the model learn fine-grained attribute disentanglement and reliable reasoning across multiple inputs. Experiments on compositional, selective and layout-based generation demonstrate consistent improvements over unified baselines such as Bagel in terms of subject fidelity, structural alignment and controllability. Ablation results further verify that each part of the framework contributes to the overall performance. SIGMA offers a model-agnostic design that can be applied to future diffusion transformers and provides a practical foundation for the study of structured conditioning and unified multi-attribute generative modeling.
\clearpage
\setcounter{page}{1}

\twocolumn[{%
\centering
{\Large\bfseries SIGMA: Selective-Interleaved Generation with Multi-Attribute Tokens \par}
\vspace{0.5em}
{\large\bfseries Appendix \par}
\vspace{1em}
}]

\appendix
\renewcommand{\thesection}{\Alph{section}}

This appendix provides additional details and analyses for SIGMA.
Section~\ref{subsec:dataset_overview_more} presents an extended description of the interleaved dataset, including task coverage, attribute design, and the construction pipeline with special tokens and interleaving strategies.
Section~\ref{sec:dataset_construction_supp} further explains how generated data and existing corpora are combined into a unified training set.
Section~\ref{sec:benchmark_supp} introduces the evaluation protocols, including XVerseBench and a comprehensive benchmark covering compositional generation, selective generation, style and relation transfer, and layout-based generation.
Section~\ref{sec:eval_train_supp} describes the evaluation and training protocols, detailing how multi-condition inputs are handled for all baselines and providing additional implementation details.
Section~\ref{sec:user_study} reports the user study setup and analyzes human preferences across different tasks, showing that SIGMA is consistently favored over strong baselines.
Section~\ref{sec:failure} presents representative failure cases and discusses the conditions under which SIGMA may struggle.
Finally, Section~\ref{sec:addition_eva} provides additional qualitative results and visual comparisons across various tasks.

\section{Dataset}
\subsection{Dataset Overview}
\label{subsec:dataset_overview_more}
To make SIGMA actually learn which attribute should be taken from which reference, we curate a large-scale, text–image interleaved dataset with explicit attribute annotations. The final corpus contains about \textbf{700K} interleaved sequences, each mixing text spans, special tokens, and one or more condition images. The data covers six major task families: \emph{compositional generation} (100K), \emph{selective extraction} (226K), \emph{style transfer} (153K), \emph{relation transfer} (41.6K), \emph{image editing} (70K), and \emph{conditional layout generation} (110K). This diversity is important: SIGMA is expected to see not only ``one prompt + one image" cases, but also multi-image, cross-attribute, and asymmetric-reference cases at training time.

Pure vision–text pairs are not enough for that goal. While vision–text paired data provides useful supervision, it falls short when the model must track multiple visual sources, resolve which text span refers to which image, or propagate an intermediate relation (e.g., ``make C look like A stylized as B"). Models trained on such data tend to blur visual and semantic correspondences across modalities, producing less coherent multi-source generations. Our dataset therefore switches the supervision unit from ``caption~$\leftrightarrow$~image" to ``interleaved sequence~$\leftrightarrow$~image": every sequence explicitly tells the model when a new visual source appears and what semantic role it plays.

To make these roles machine-readable, we introduce 14 special tokens that indicate specific visual entities or attributes (e.g., identity, subject, style, clothing, background, layout, pose, lighting, texture, emotion), together with a small functional token for explicit extraction of designated objects. Instead of describing ``a red bag next to the person" only in free-form text, we insert the corresponding token right before the textual mention and then place the referenced image in the sequence. This tokenization does two things at once: it anchors the text span to the correct image group, and it disambiguates which attribute we want to pull from that image. In compositional generation, where several reference images may contain overlapping content or similar colors, this explicit anchoring is crucial. Otherwise, the model easily drifts and copies the wrong source.

Another key design choice is that the dataset is intentionally attribute dense.
Many samples contain multiple attribute tokens within the same image, for example \texttt{<subject>} \texttt{<clothing>} and \texttt{<background>} appearing together.
This reflects the realistic setting of interleaved generation because when users provide several reference images, the ambiguity of what should be transferred increases very quickly.
Training on such densely annotated and multi attribute cases helps SIGMA learn to extract only the factors that are explicitly requested by the instruction and to take them from the correct reference image rather than mixing all visual cues together.
Together with the scale of the dataset, which contains 700K samples, and the diversity of included tasks, this design enables SIGMA to maintain strong controllability at inference time even when the input conditions are heterogeneous and user specified.

\subsection{Dataset Construction}
\label{sec:dataset_construction_supp}

\paragraph{Data Sources and Generation.}
Our interleaved dataset combines both newly generated and adapted sources to ensure broad coverage of compositional and conditional generation scenarios.
We create large-scale synthetic data for compositional, stylization, and editing tasks, and further incorporate high-quality existing corpora augmented with explicit image–token bindings.

We generate about 220K compositional samples using GPT-4o~\cite{openai2024gpt4ocard} and Nano-Banana~\cite{comanici2025gemini25pushingfrontier}, covering diverse human–object–scene combinations such as indoor/outdoor environments, product displays, and clothing replacement.
A 30K stylization subset and a 41.6K style-relation transfer subset capture style and relation mappings; the latter trains the model to apply a relation transformation $(A \rightarrow B)$ to a new content image $C$, with style references collected from PromptsRef and de-styled variants produced by Gemini~2.5~Pro.
We additionally include 20K image-editing samples generated with Nano-Banana for customized content replacement and object insertion, verified for visual coherence.
For selective-content extraction, we adapt Echo-4o~\cite{ye2025echo4oharnessingpowergpt4o} by treating each compositional output as input and identifying target objects via GPT-4o analysis, producing prompts such as “Extract the \{object\} from the image.”
The conditional-layout subset includes two types of samples: those using explicit layout inputs such as canny or depth maps, and those where one image from a compositional sequence is designated as the layout condition.
Depth and structural cues are extracted using MiDaS~\cite{ranftl2020towards}, ensuring consistent geometric control across diverse scenes.

Existing datasets including Nano-150K~\cite{ye2025echo4o}, Echo-4o~\cite{ye2025echo4oharnessingpowergpt4o}, X2Edit~\cite{ma2025x2editrevisitingarbitraryinstructionimage}, and ShareGPT-4o~\cite{chen2025sharegpt4oimagealigningmultimodalmodels} are also integrated after token-based alignment, forming a unified interleaved corpus for multi-condition learning.

\paragraph{Interleaving and Token Injection Pipeline}
\label{subsubsec:interleave_pipeline}
To unify data from diverse sources under the interleaved paradigm, we design a structured token-injection pipeline that transforms each original text–image pair into an interleaved multimodal sequence.
Given a caption describing multiple entities or attributes, we first parse the sentence to locate phrases referring to concrete visual elements (e.g., ``the man," ``the red car," ``the city background," ``the jacket").
Each identified phrase is prepended with a corresponding \texttt{<special\_token>}, such as \texttt{<id>}, \texttt{<subject>}, \texttt{<clothing>}, or \texttt{<background>}, which specifies the intended attribute.
The referenced image is then inserted immediately after the phrase, forming a locally bound text–image pair. Formally, a standard caption such as ``Make the man wear the blue jacket and stand beside the car." is converted into 
``Make \texttt{<id>}the man in \texttt{<}portrait image\texttt{>} wear \texttt{<clothing>}the blue jacket in \texttt{<}jacket image\texttt{>} and stand beside \texttt{<subject>}the car in \texttt{<}car image\texttt{>}"
This procedure yields an interleaved sequence that the diffusion transformer can process directly, with each token–image group representing one condition scope.

For existing datasets without explicit entity information, we employ a large-language-model parser to infer likely correspondences between nouns and reference images.
Each inferred entity is assigned an appropriate token from the 14-token vocabulary, ensuring consistent semantics across heterogeneous data sources.
We further enforce quality control by discarding cases with ambiguous references or unresolved entities.

This standardized interleaving pipeline transforms traditional vision–language datasets into structured multimodal supervision that explicitly encodes cross-condition relationships.
By training on such sequences, SIGMA learns to associate token semantics with their corresponding visual sources, enabling robust multi-attribute reasoning and fine-grained controllability during generation.

\section{Benchmarks}
\label{sec:benchmark_supp}
We evaluate our method and baselines and cover different aspects of controllable image generation.
XVerseBench~\cite{chen2025xverse} is used for compositional generation, comprising 20 distinct human identities, 74 unique objects, and 45 different animal species or individuals.
The compositional generation split contains 210 cases, each involving 1--3 input combinations, allowing us to evaluate the model's ability to handle multi-entity reasoning.

We further introduce a new \textbf{comprehensive benchmark} that jointly evaluates controllability, compositional reasoning, and structural alignment across four representative tasks:
\emph{compositional generation}, \emph{selective generation}, \emph{style transfer}, and \emph{layout-based generation}.
All benchmark samples are constructed from the held-out portion of our corpus that is never used for training, ensuring a strict separation between training and evaluation data. 
Specifically, the compositional subset contains 100 samples with 2 to 6 inputs that mix humans, objects, and scenes. The selective subset includes 100 samples that focus on extracting or generating specific subjects. The style transfer task contains 60 samples, and the relation transfer task also contains 60 samples, each designed to evaluate a different aspect of style/relation control. The layout based subset includes 70 samples that are guided by edge based conditions, where 17 use a single canny map and 53 combine canny maps with additional reference images. 

Together, XVerseBench and our comprehensive benchmark provide complementary perspectives. The former serves as a public and standardized evaluation protocol, while the latter delivers a broad and fine-grained assessment of controllability across diverse multi-condition settings. These two benchmarks ensure that the evaluation captures both external benchmark performance and generalization across rich, realistic, and systematically varied generation scenarios.

\begin{figure*}[t]
    \centering
    \begin{subfigure}[b]{0.32\textwidth}
        \centering
        \includegraphics[width=\linewidth]{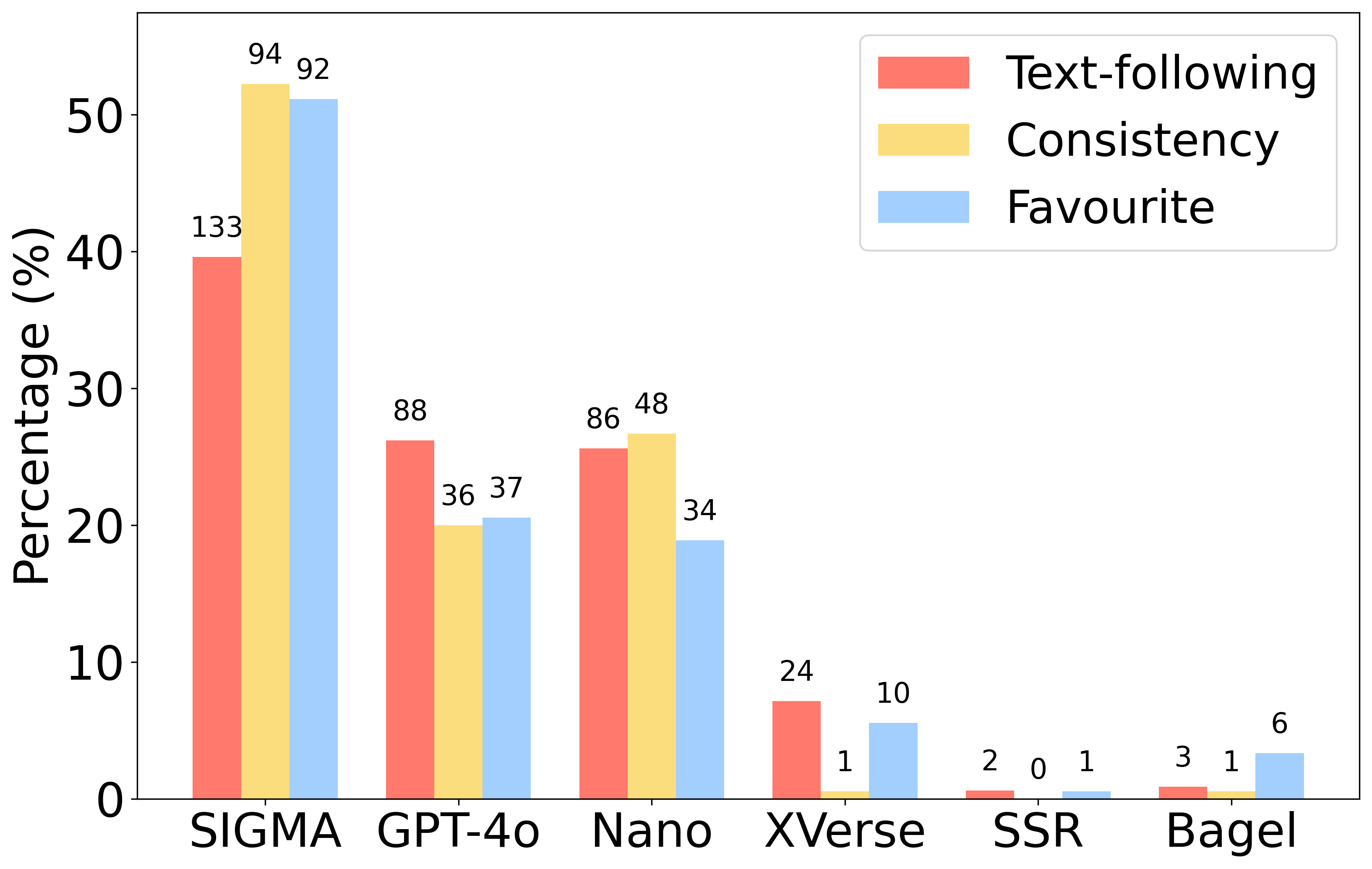}
        \caption{Compositional generation}
        \label{fig:user-study-comp}
    \end{subfigure}
    \hfill
    \begin{subfigure}[b]{0.32\textwidth}
        \centering
        \includegraphics[width=\linewidth]{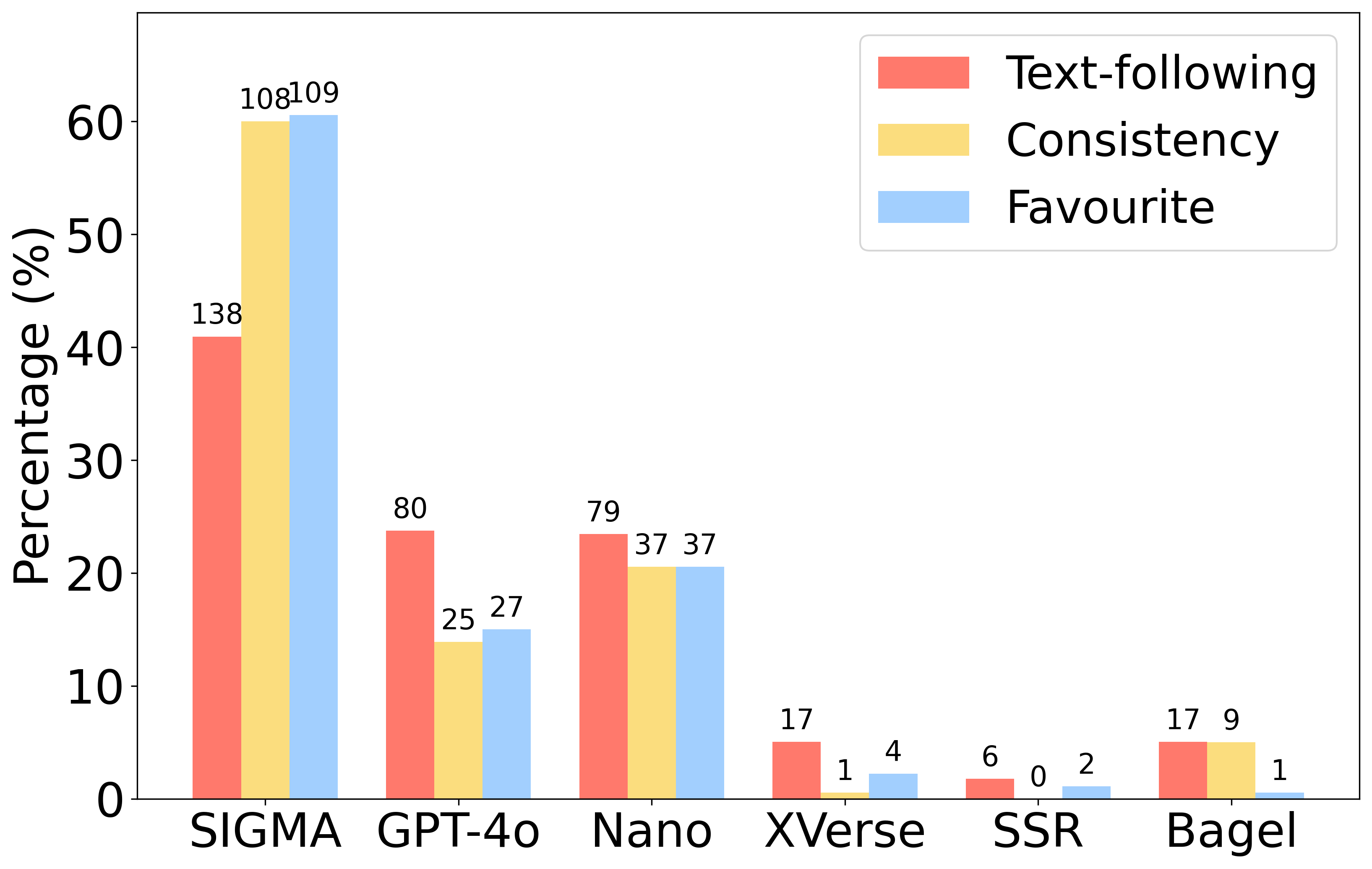}
        \caption{Selective generation}
        \label{fig:user-study-sel}
    \end{subfigure}
    \hfill
    \begin{subfigure}[b]{0.32\textwidth}
        \centering
        \includegraphics[width=\linewidth]{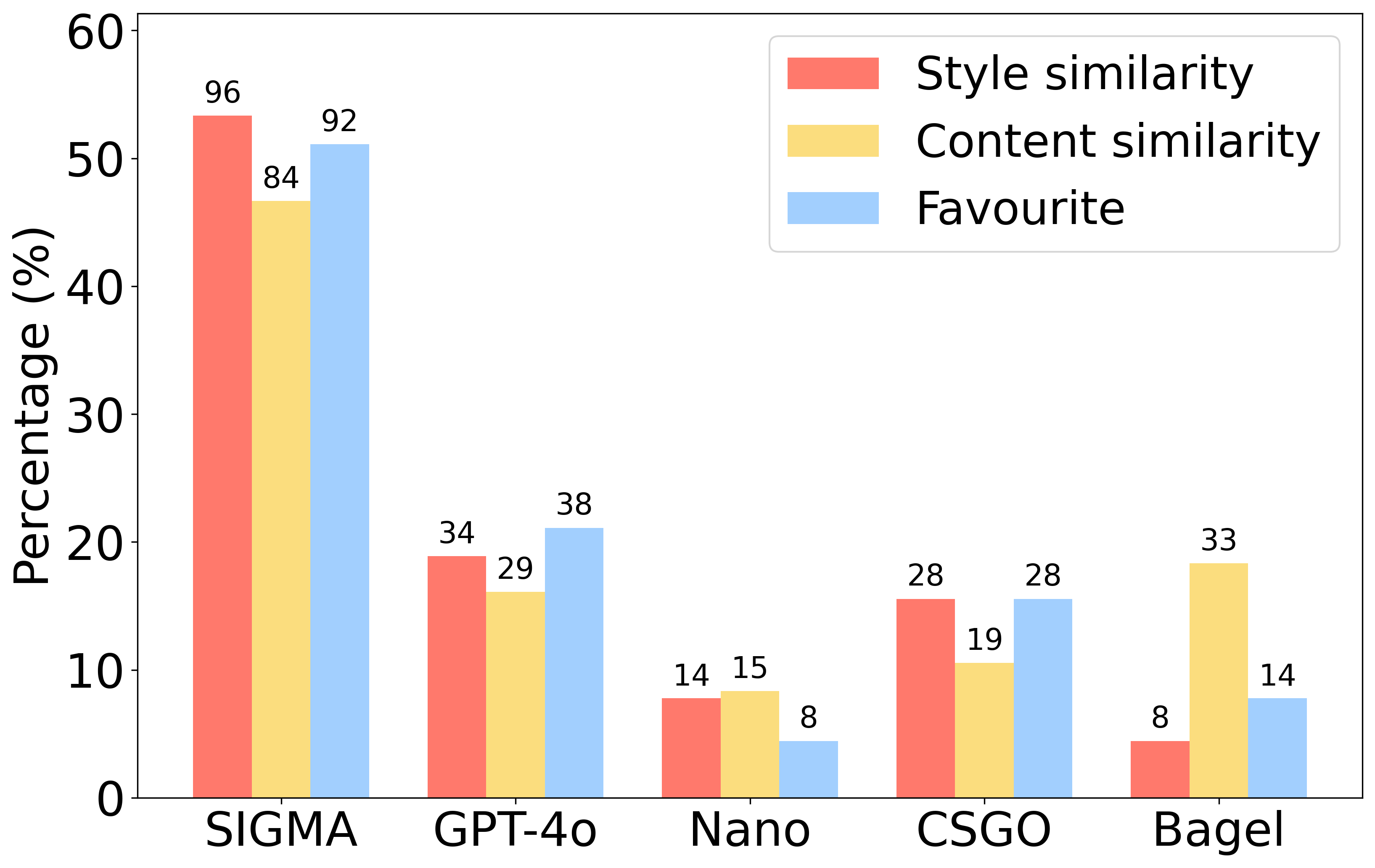}
        \caption{Style transfer}
        \label{fig:user-study-style}
    \end{subfigure}

    \caption{User study preference across the three tasks.}
    \label{fig:user-study-overall}
\end{figure*}

\section{Evaluation and Training Protocols}
\label{sec:eval_train_supp}
\subsection{Evaluation Protocol for Baselines}
To ensure a fair comparison across all methods, we provide each baseline with multi-condition inputs in the format natively supported by that model, without altering its original conditioning mechanism.

For both SIGMA and the Bagel~\cite{deng2025emergingpropertiesunifiedmultimodal}, inputs follow the text–image interleaving format used during post-training. Each reference image is inserted as an image token block, and SIGMA additionally prepends the corresponding special attribute token to explicitly specify the role of each reference. Bagel receives the same interleaved sequence but without attribute tokens, resulting in a purely text–image alternating structure. For SSR~\cite{zhang2024ssr} and XVerse~\cite{chen2025xverse}, each reference image is provided together with an associated subject identifier, following the native conditioning design of these models.
For GPT-4o~\cite{openai2024gpt4ocard} and Nano~\cite{comanici2025gemini25pushingfrontier}, only standard text and image inputs are supported. Since neither model accepts interleaved multimodal sequences or explicit attribute tokens, we provide the full textual instruction followed by all reference images in a flat, unordered format. Generation is then performed through a single forward pass without any model-specific tuning or adaptation. This protocol evaluates each model strictly under its native interface, ensuring that differences in performance arise from controllability rather than input formatting constraints.

\subsection{Extended Implementation Details}
All images are processed at the native spatial resolution of the Bagel backbone. We employ the backbone’s default diffusion sampler and scheduling strategy, and use classifier-free guidance with a fixed guidance scale throughout training and evaluation. During both training and inference, no test-time optimization, prompt rewriting, or auxiliary adapters are used. For all baselines, we directly follow their original sampling hyperparameters and native inference settings to avoid introducing differences unrelated to model capability.

\section{User Study}
\label{sec:user_study}
To complement automatic metrics, we conduct a user study to assess the perceptual quality of SIGMA in compositional generation, selective generation, and style transfer.
\subsection{User Study Setup}
\paragraph{Task design.}
We construct a user study with a total of 15 groups of samples.
Each of the three tasks contains 5 samples in total.  
For each sample, the images produced by different methods are anonymized and randomly shuffled before being shown to users.
The compositional and selective generation tasks include 5 additional methods in total, namely, GPT-4o~\cite{openai2024gpt4ocard}, Nano~\cite{comanici2025gemini25pushingfrontier}, XVerse~\cite{chen2025xverse}, SSR~\cite{zhang2024ssr}, and Bagel~\cite{deng2025emergingpropertiesunifiedmultimodal}, while the style-transfer task evaluates 4 additional methods, with CSGO~\cite{xing2024csgo} replacing XVerse and SSR.
Each sample is associated with three multiple-choice questions, leading to 45 question blocks in total.
All tasks share a common question that asks users to select their \emph{overall favorite} image among all candidates.
For compositional generation, the remaining two questions are:
(i) ``Select the result that follows the textual instruction,'' which evaluates text-following ability; and
(ii) ``Select the result where the objects are most visually consistent with the corresponding objects in the input images,'' which evaluates visual consistency and correct attribute binding.
For selective generation, we use the same two task-specific questions as compositional generation to access the ability of text following and keeping consistency respectively.
For style transfer, the two task-specific questions are:
(i) ``Select the result whose \emph{content} is most similar to the content input image,'' which measures content preservation; and
(ii) ``Select the result whose \emph{style} is most similar to the style input image,'' which measures style similarity.
The text-following questions allow users to select more than one result when several outputs follow the instruction equally well. In all other cases, users provide a single choice per question.

\paragraph{Participants.}
We collected 36 valid responses in total after discarding incomplete questionnaires.
The participants include both researchers and university students, covering people with prior experience in computer vision and image generation as well as non-expert users.
Each participant answered all 15 samples, resulting in $36 \times 15 \times 3$ individual choices overall.

% ...

\subsection{Results and Analysis}
To quantify human preference across different aspects of generation quality, we aggregate user votes by task, question type, and method.
For each task, we sum all votes belonging to the same evaluation question (instruction following, consistency, or style/content similarity), and normalize the counts to obtain preference percentages for each competing method.
Figure~\ref{fig:user-study-overall} visualizes the results for the three tasks.
Within each subplot, bars of the same color correspond to the same evaluation aspect, the bar height indicates the normalized preference, and the number above each bar denotes the raw vote count.

As shown in Figure~\ref{fig:user-study-comp} and Figure~\ref{fig:user-study-sel}, SIGMA receives the highest user preference in both compositional generation and selective generation across all evaluation aspects.
In particular, SIGMA achieves the largest proportion of instruction-following votes and a markedly higher proportion of consistency votes than all baselines, indicating that users perceive SIGMA as better at preserving object identity and maintaining correct attribute binding during generation.
The consistently strong performance on these two tasks suggests that SIGMA more reliably extracts object-level cues from input references and integrates them into the final images in a faithful and coherent way.
Figure~\ref{fig:user-study-style} shows that SIGMA also performs strongly in the style-transfer task.
SIGMA attains the highest preference in both style similarity and content similarity, demonstrating its ability to model complex style cues while avoiding conflicts between transferred style and preserved content.
The strong user preference in the ``favourite" question further suggests that SIGMA produces visually well-balanced results that users find both natural and aesthetically appealing.


\section{Failure Cases}
\label{sec:failure}
\begin{figure}[t]
    \centering
    \includegraphics[width=1.0\linewidth]{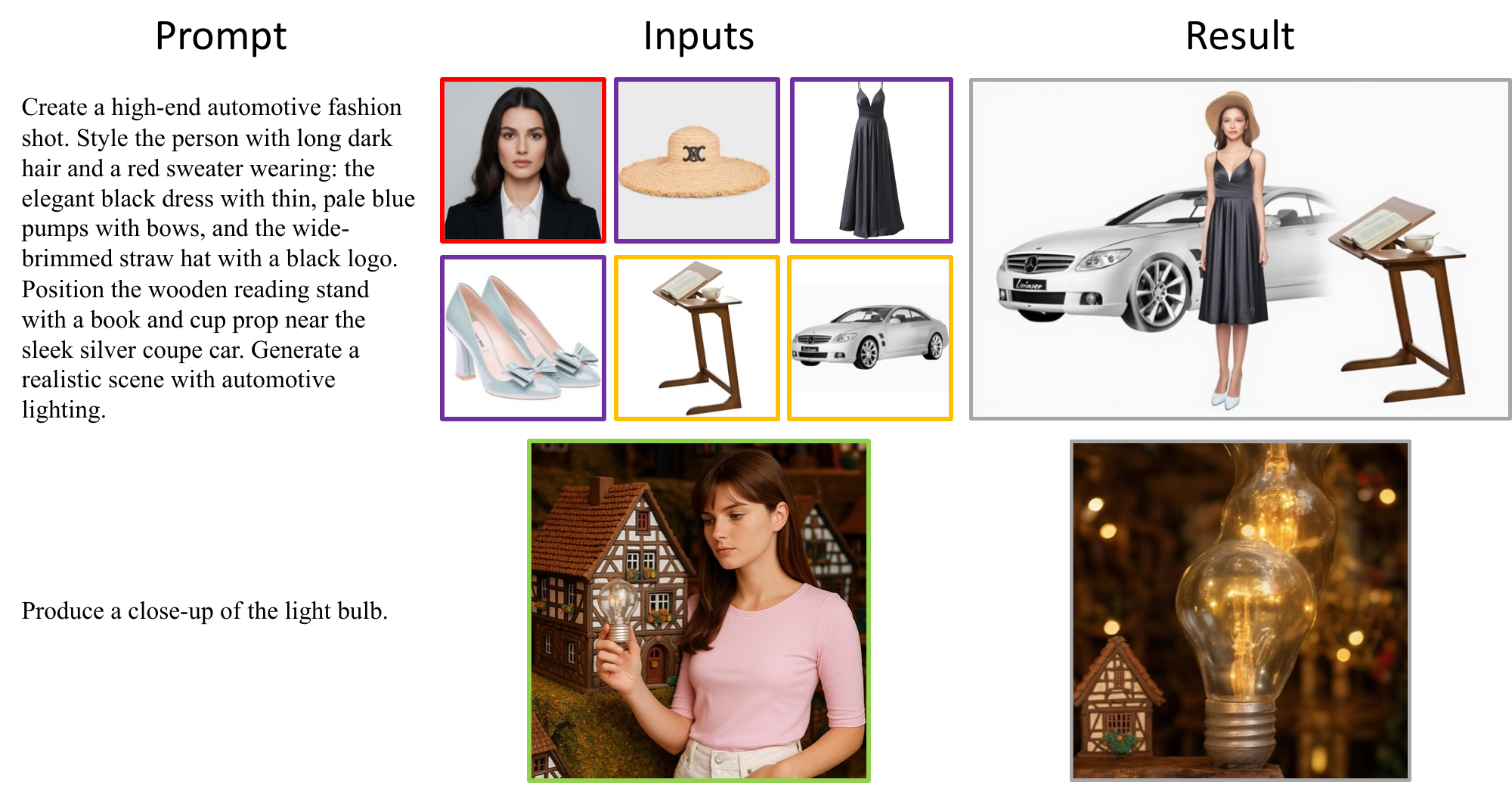}
    \caption{Two representative failure cases of SIGMA. Top: a compositional generation example with a large number of heterogeneous references. Bottom: a selective generation case where the reference image contains complex background structures.}
    \label{fig:failure}
\end{figure}

Figure~\ref{fig:failure} illustrates two typical failure cases encountered in our framework, covering compositional generation and selective extraction.
The first example is a high-complexity compositional generation task. The prompt requires simultaneously binding multiple heterogeneous references, including a person, hat, dress, shoes, a reading stand, and a vehicle. When the number of input entities becomes large, the model occasionally struggles to maintain natural interactions across all referenced objects. This results in less coherent spatial arrangements as well as incomplete rendering of certain elements, such as the partially blurred rear of the car. These issues reflect the challenge of handling dense multi-entity reasoning when the reference set grows beyond the typical range seen during training.
The second example shows a selective generation failure, where the model is asked to isolate and render a specific object from the reference image containing a complex, cluttered background. When the target object is small relative to its surroundings and embedded within highly textured environments, the attribute-specific token fails to fully suppress distractors, leading to geometric deformation or leakage of irrelevant background features. This indicates that selective extraction becomes more difficult when the foreground object is visually entangled with the background, particularly in cases where object boundaries are not sharply separable.

These cases highlight limitations that arise mainly in extreme or visually challenging scenarios, either when the number of referenced entities exceeds the model’s typical operating regime or when selective extraction is applied to cluttered scenes with non-salient targets. Improving robustness under these conditions remains an interesting direction for future works.

\section{Additional Qualitative Evaluation}
\label{sec:addition_eva}
\begin{figure*}[t]
    \centering
    \includegraphics[width=\linewidth]{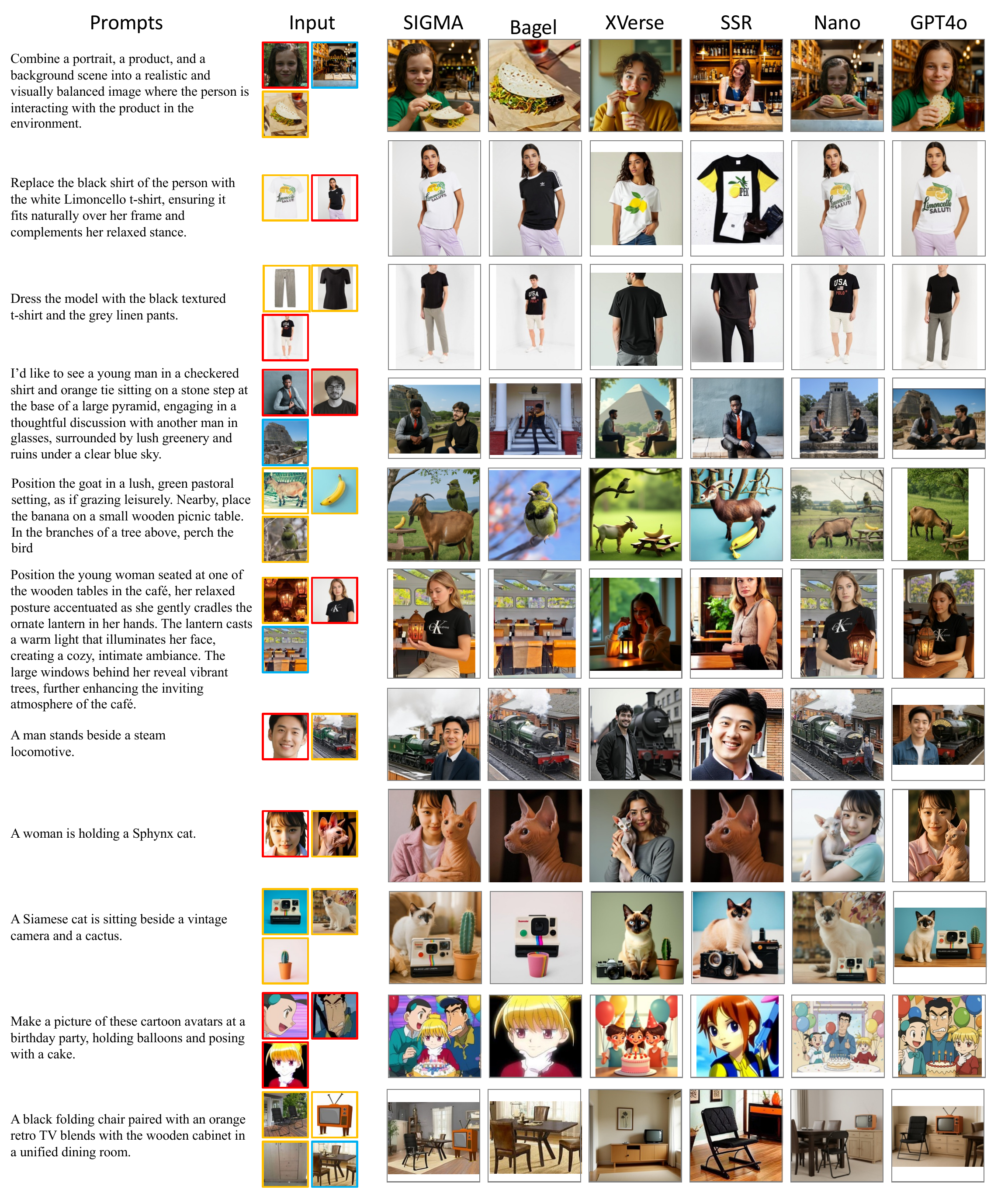}
    \caption{Additional compositional generation results.  
    SIGMA preserves object identity, maintains attribute binding across references, and produces coherent global structures, while baselines often struggle with object fusion, spatial arrangement, or cross-object consistency.
    }
    \label{fig:more_composition}
\end{figure*}

\begin{figure*}[t]
    \centering
    \includegraphics[width=\linewidth]{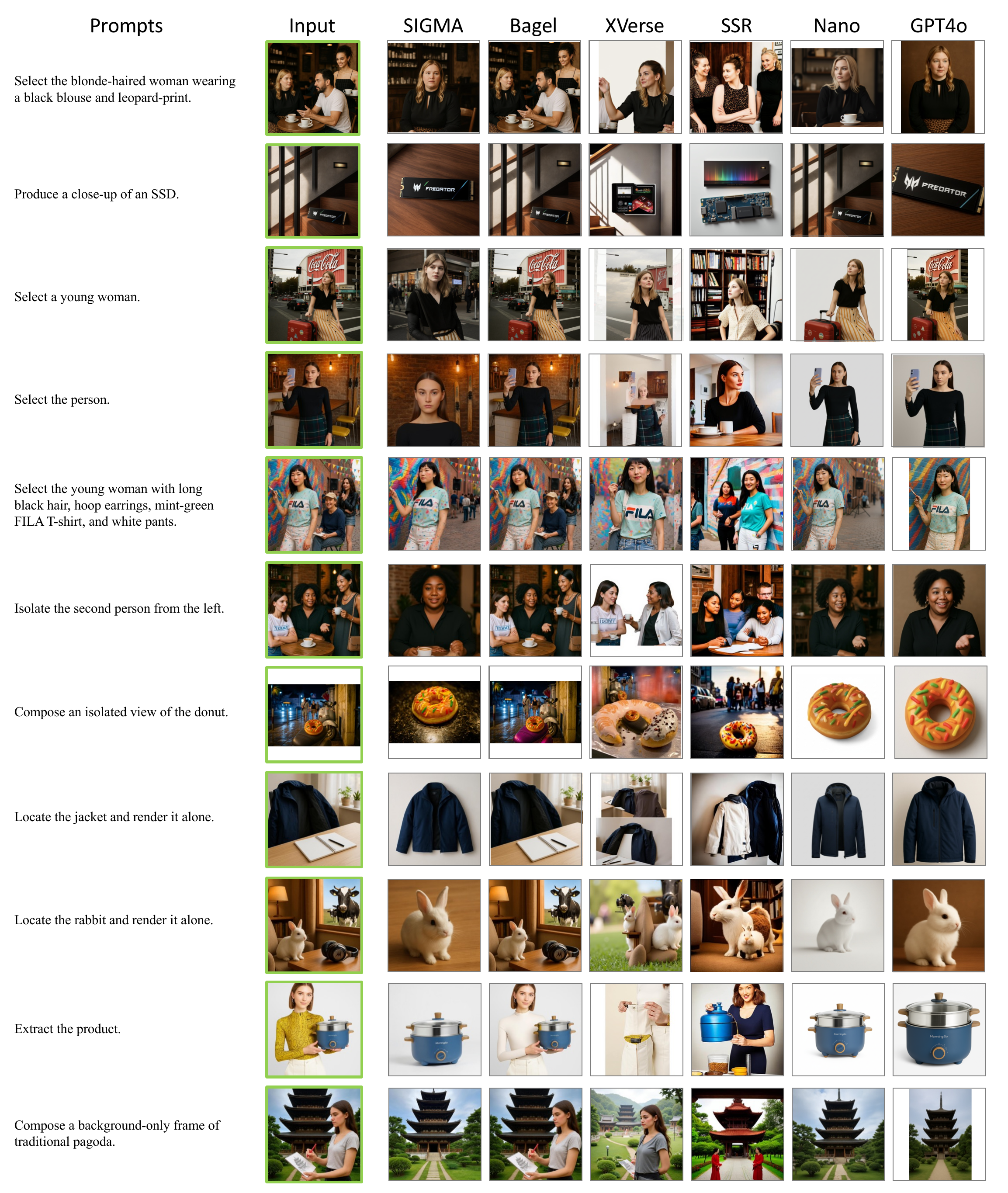}
    \caption{Additional selective generation examples.  
    Given an instruction that targets only part of the scene, SIGMA reliably extract the specified region while keeping the appearance unchanged.  
    Compared with the baselines, SIGMA achieves stronger localized control with fewer artifacts and better structural preservation.}
    \label{fig:more_select}
\end{figure*}

\begin{figure*}[t]
    \centering
    \includegraphics[width=\linewidth]{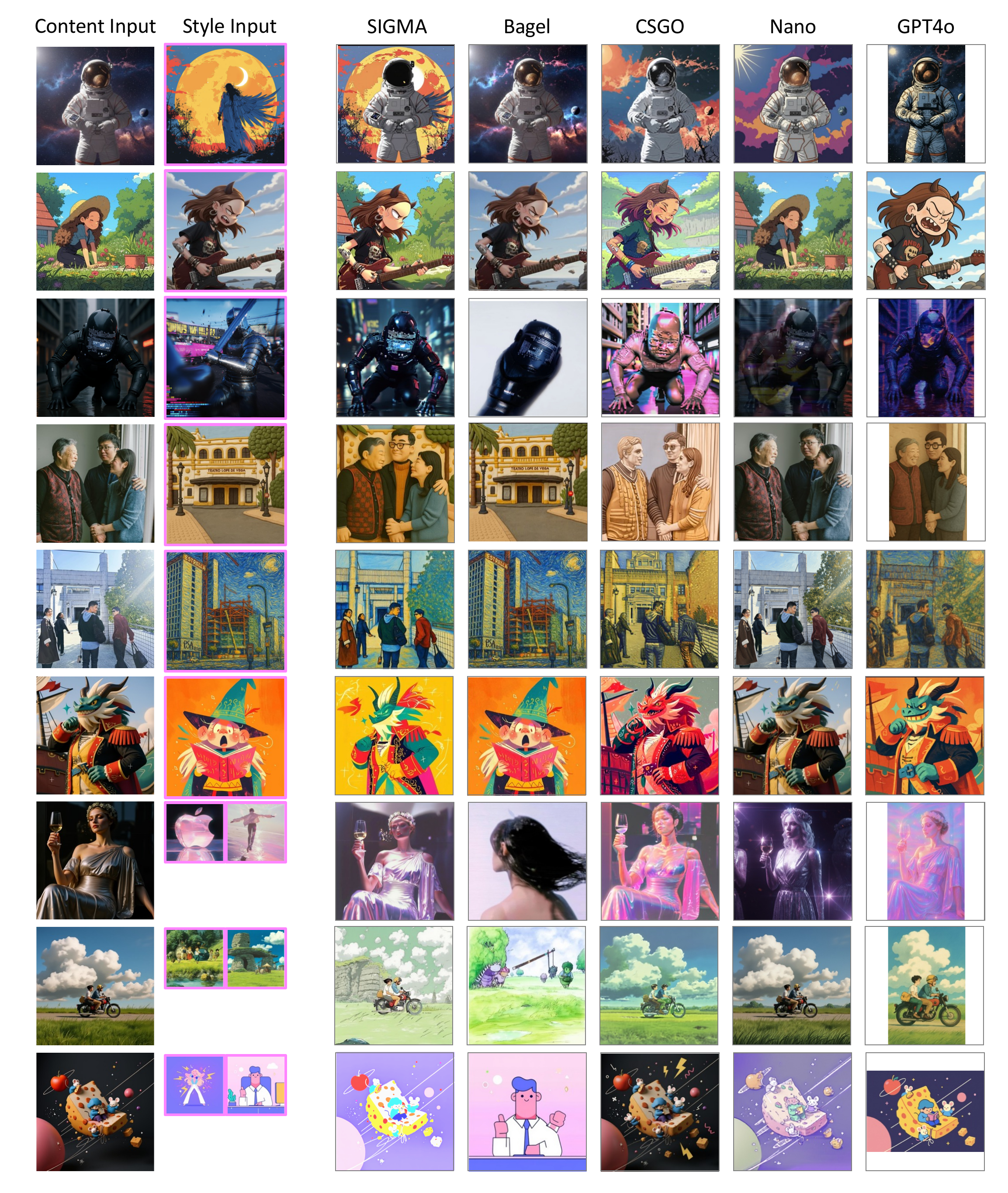}
    \caption{Additional style-transfer comparisons.  
    SIGMA successfully transfers fine-grained style cues from the reference image onto the content image while maintaining structure and detail.  
    }
    \label{fig:more_style_transfer}
\end{figure*}

\begin{figure*}[t]
    \centering
    \includegraphics[width=\linewidth]{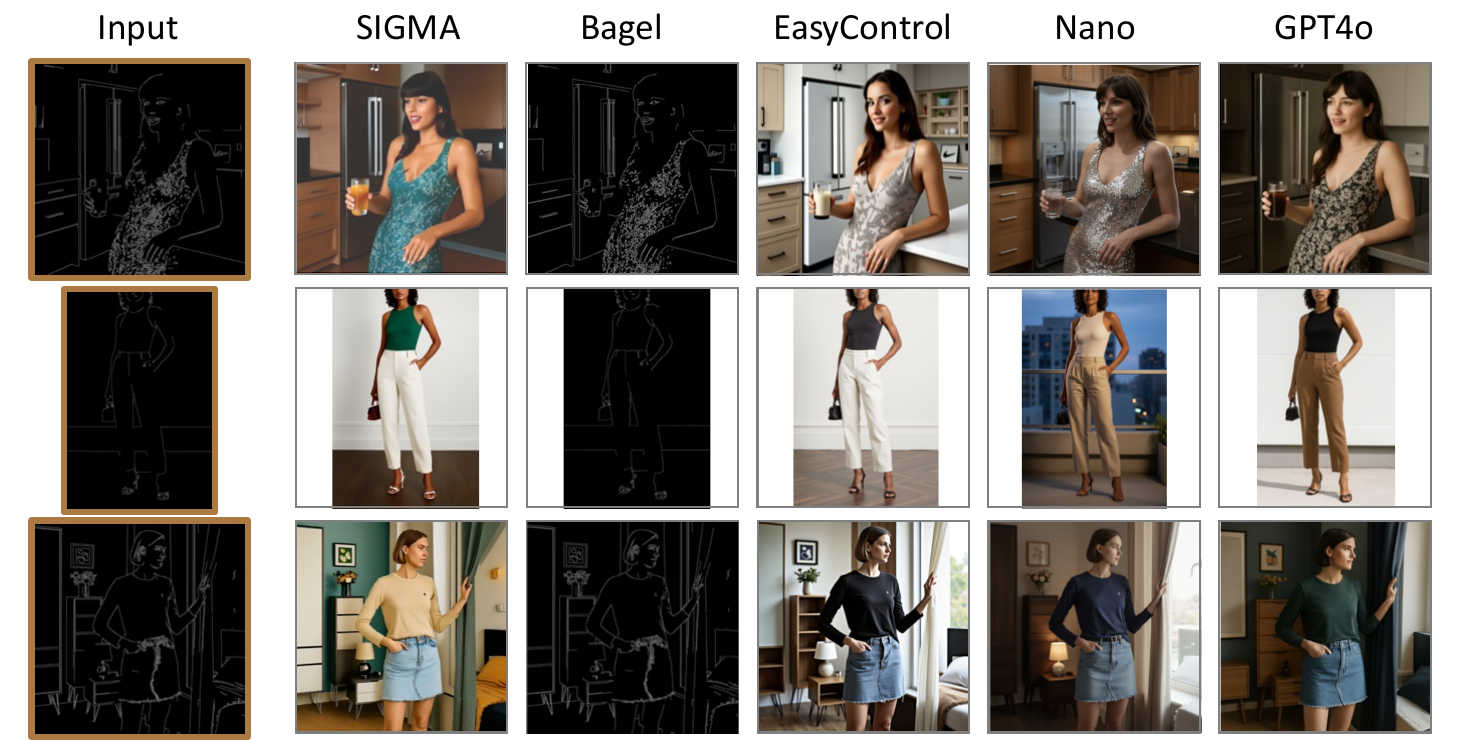}
    \caption{Additional layout-only generation results.  
    Given only a spatial layout as conditioning, SIGMA adheres closely to the prescribed structure while generating high-quality and coherent content.}
    \label{fig:more_layout_only}
\end{figure*}

\begin{figure*}[t]
    \centering
    \includegraphics[width=\linewidth]{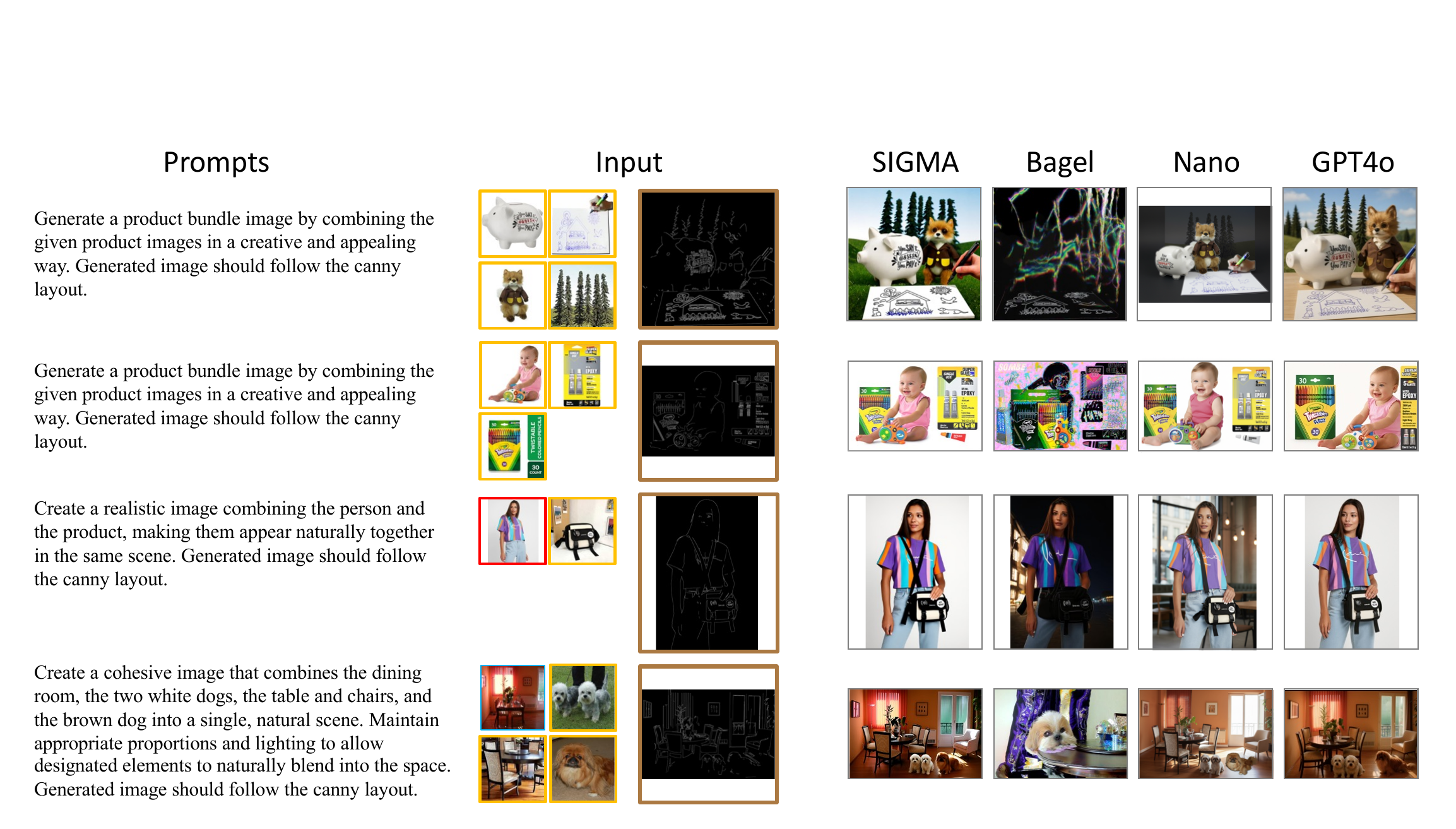}
    \caption{Additional layout + reference generation examples.  
    SIGMA effectively combines layout constraints with appearance cues from the reference image, producing results that align with both spatial structure and visual identity.  
    Baselines exhibit layout drift or fail to transfer reference-specific details, while SIGMA maintains precise structure and faithful attribute transfer.}
    \label{fig:more_layout_reference}
\end{figure*}

\begin{figure*}[t]
    \centering
    \includegraphics[width=\linewidth]{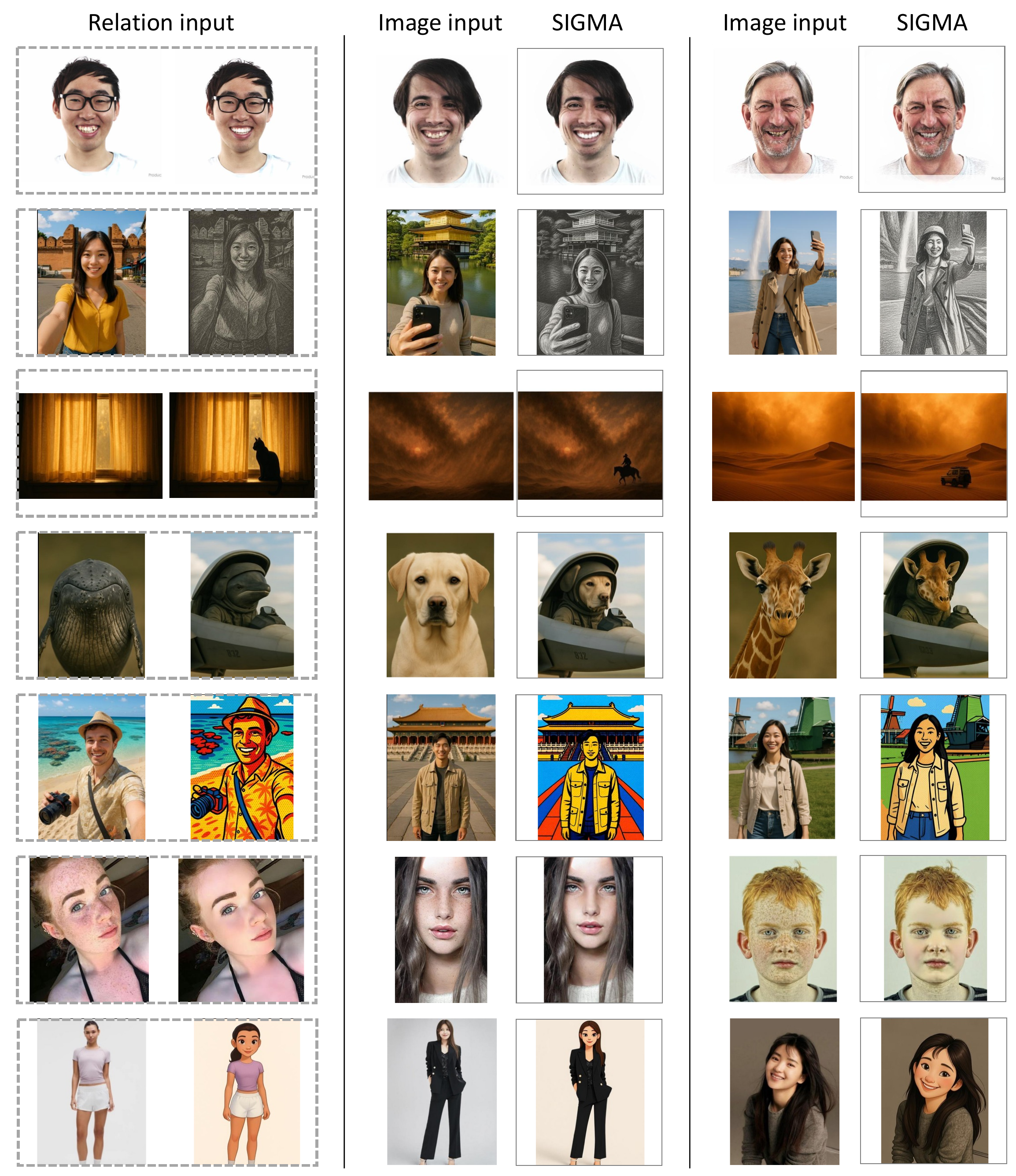}
    \caption{Additional relation-transfer results.  
    SIGMA reproduces the relational configuration between reference objects, such as relative pose, interaction, or style, while allowing appearance variations.  
    These examples demonstrate SIGMA’s ability to capture higher-order visual relationships beyond object-level cues.}
    \label{fig:more_relation_transfer}
\end{figure*}

\begin{figure*}[t]
    \centering
    \includegraphics[width=\linewidth]{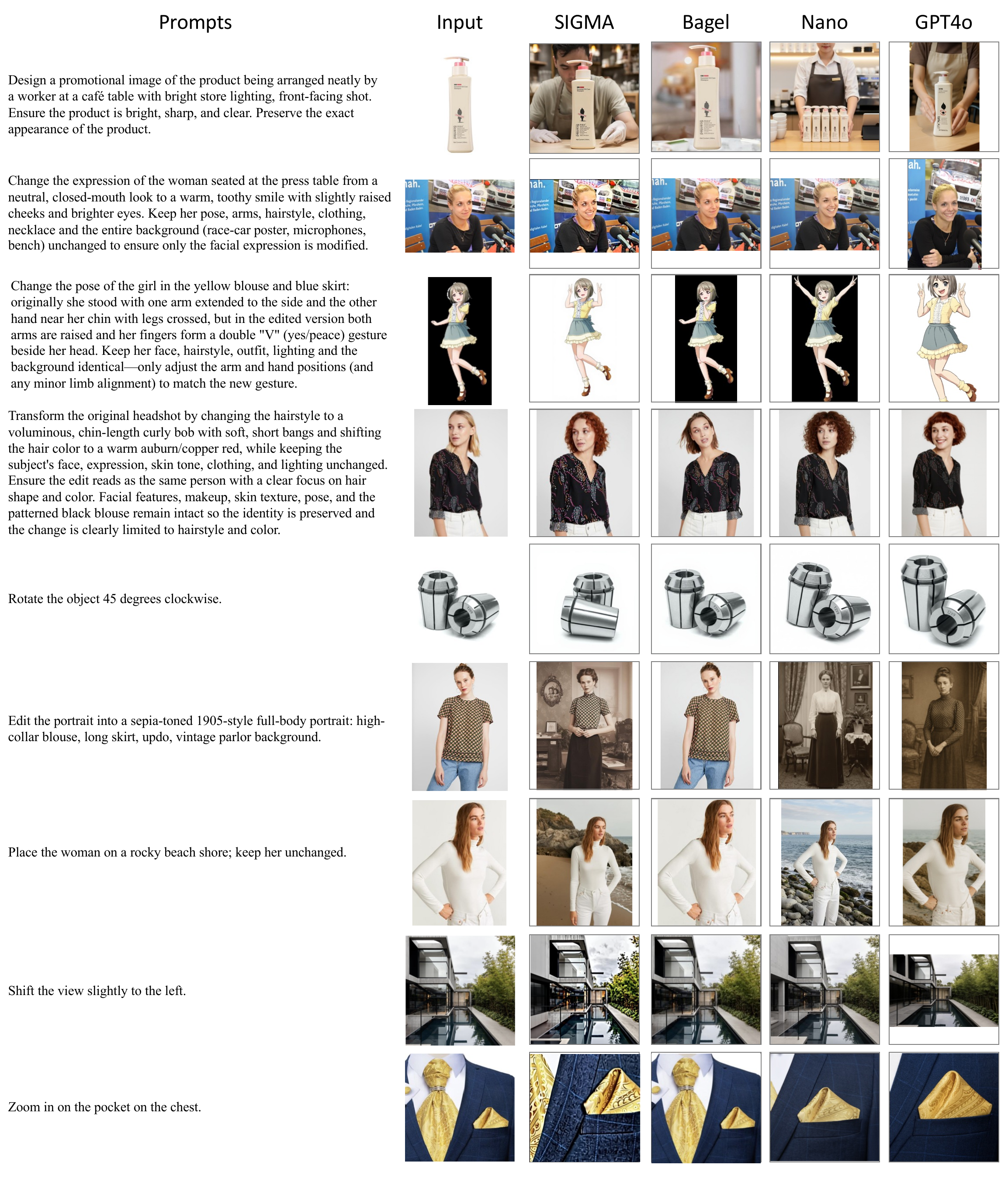}
    \caption{Image-editing results.  
    SIGMA modifies the requested content according to the text prompts while preserving surrounding regions, global scene layout, and overall visual coherence.  
    }
    \label{fig:more_edit}
\end{figure*}

We further provide an extensive set of qualitative results to illustrate the generality and robustness of SIGMA across a wide range of generation scenarios. For compositional generation, Figure~\ref{fig:more_composition} shows additional comparisons against all baselines, where SIGMA consistently produces coherent combinations that preserve object identity and correctly bind attributes across references. In selective generation (Figure~\ref{fig:more_select}), SIGMA reliably extracts only the instructed regions or objects while maintaining the appearance and structure. Additional style-transfer results in Figure~\ref{fig:more_style_transfer} show that SIGMA captures complex stylistic cues while preserving content, producing clean and visually stable outputs. We also include layout-only generation results in Figure~\ref{fig:more_layout_only}, where SIGMA adheres faithfully to coarse spatial layouts while generating high-quality content, as well as layout + reference generation in Figure~\ref{fig:more_layout_reference}. Relation-transfer examples are provided in Figure~\ref{fig:more_relation_transfer}, illustrating SIGMA’s ability to reproduce relational configurations such as relative poses or style. Finally, Figure~\ref{fig:more_edit} presents editing examples, showing that SIGMA performs competitively on diverse edit instructions while preserving overall scene fidelity. These extended qualitative results highlight SIGMA’s versatility, its strong attribute and relation binding, and its overall stability and fidelity across a broad spectrum of generation and editing tasks.

\vspace{20mm}

{
    \small
    \bibliographystyle{ieeenat_fullname}
    \bibliography{main}
}

\end{document}